\documentclass{article}
\usepackage{iclr2027_conference,times}
\AtBeginDocument{\DeclareFontShape{OT1}{ptm}{m}{scit}{<->ssub*ptm/m/sc}{}}

\usepackage{amsmath,amsfonts,bm}

\def\eqref#1{equation~\ref{#1}}

\def\1{\bm{1}}

\DeclareMathAlphabet{\mathsfit}{\encodingdefault}{\sfdefault}{m}{sl} \SetMathAlphabet{\mathsfit}{bold}{\encodingdefault}{\sfdefault}{bx}{n}

\usepackage[hidelinks]{hyperref}
\usepackage{url}
\usepackage{booktabs}
\usepackage{graphicx}
\usepackage{wrapfig}
\usepackage{placeins}
\usepackage[table]{xcolor}
\usepackage{capt-of}
\usepackage{etoc}
\usepackage{multirow}
\makeatletter
\def\section{\@startsection{section}{1}{\z@}{-1.6ex plus -0.4ex minus -.2ex}%
{0.9ex plus 0.2ex}{\large\sc\raggedright}}
\def\subsection{\@startsection{subsection}{2}{\z@}{-1.3ex plus -0.4ex minus -.2ex}%
{0.6ex plus .2ex}{\normalsize\sc\raggedright}}
\def\paragraph{\@startsection{paragraph}{4}{\z@}{1.5ex plus
0.1ex minus .2ex}{-1em}{\normalsize\bf}}
\makeatother

\newcommand{\repourl}{https://github.com/HaohanYuan01/ForgePrint}

\newcommand{\method}{\textsc{ForgePrint}}
\graphicspath{{figures/pdf/}}

\providecommand{\blk}[1]{\textit{\textcolor{black!70}{#1}}}

\providecommand{\dn}[1]{\raisebox{-0.75ex}{\tiny\textcolor{dropred}{$-$#1}}}
\definecolor{dropred}{HTML}{B3352B}
\providecommand{\ci}[1]{{\tiny\textcolor{cigrey}{$\pm$#1}}}
\definecolor{cigrey}{HTML}{8A8A8A}
\definecolor{ourrow}{HTML}{E8F0F9}

\usepackage[most]{tcolorbox}
\definecolor{cUnmod}{HTML}{E6E6E6}   
\definecolor{cBase}{HTML}{BDBDBD}    
\definecolor{cTeach}{HTML}{6BAED6}   
\definecolor{cOurs}{HTML}{4292C6}    
\definecolor{cTarget}{HTML}{FBE1C8}  
\definecolor{cCue}{HTML}{D95F02}     
\newcommand{\cue}[1]{\textcolor{cCue}{#1}}
\newtcolorbox{casecard}[2][]{%
  breakable, enhanced, before skip=8pt, after skip=8pt,
  colback=white, colframe=black!55, boxrule=0.4pt, arc=1.5pt,
  left=5pt, right=5pt, top=4pt, bottom=4pt,
  fontupper=\footnotesize, fonttitle=\footnotesize\bfseries, coltitle=black,
  colbacktitle=black!7, title={#2}, #1}
\newcommand{\verdictline}[1]{{\scriptsize\hspace{0.8em}evaluators: #1}}
\newcommand{\roleline}[3]{\par\medskip\noindent\colorbox{#1}{\scriptsize\bfseries\strut #2}#3\par\nobreak\smallskip\noindent\ignorespaces}

\title{Forging LLM Authorship Fingerprints \\ with Targeted Rewriting}

\author{%
\small Haohan Yuan$^{1}$ \quad Simin Chen$^{2}$ \quad Xi Niu$^{1}$ \quad Hanqing Guo$^{3}$ \quad Depeng Xu$^{1}$ \quad Haopeng Zhang$^{1}$\thanks{Corresponding author.} \\[2pt]
\small $^{1}$University of North Carolina at Charlotte \quad $^{2}$George Mason University \quad $^{3}$Indiana University}

\iclrfinalcopy

\begin{document}

\maketitle
\lhead{}
\renewcommand{\headrulewidth}{0pt}
\etocsettocdepth.toc{none}


\begin{abstract}
Model-attribution classifiers can often identify which language model
produced a text, making model-specific writing patterns a  signal
of provenance.
Accurate attribution on unmodified text, however, does not show whether
the prediction still identifies the original source after deliberate rewriting.
We formulate this problem  as \emph{targeted fingerprint transfer}: rewriting one
model's output so that attribution classifiers assign it to a chosen target
model.
We study summarization, where different models receive the same document
and express the same underlying content, providing a controlled setting for
conditional generation.
We introduce \method{}, a search-then-distil framework that first searches for
rewrites that move attribution toward a target fingerprint, then distils
the selected rewrites into a one-pass $4$B Student model.
On CNN/DM, the Student reaches $70.2\%$ target success rate, outperforming both
its Teacher ($54.1\%$) and the strongest of six published rewriting baselines
($39.3\%$), against held-out classifiers that are never queried by the attack. 
It also reaches $68.3\%$ target success when transferring summaries from an
open model toward chosen commercial models.
These results show that fingerprint detectability should not be conflated with source authenticity, and that text-only attribution can provide misleading evidence of model identity under targeted rewriting,
even when it is accurate on unmodified text.
Our code and evaluation suite will be released at \url{\repourl}.
\end{abstract}

\section{Introduction}
\label{sec:intro}

Language models leave model-specific patterns in generated text,
allowing classifiers to identify the source model with high accuracy
\citep{uchendu2020authorship,sun2025idiosyncrasies}.
We refer to these patterns as \emph{authorship fingerprints}.
Such fingerprints can serve as text-based provenance signals, for example
when auditing whether a service substitutes a cheaper model for a claimed
commercial model \citep{gao2025model,cai2025getting}, or presents another
model's outputs as its own
\citep{anthropic2026threatreport}.
Accurate attribution on unmodified text, however, does not show whether
attribution remains tied to the source model after deliberate rewriting.

Existing work has mainly studied source-model attribution on unmodified
or generically rewritten text, while attacks on AI-text and authorship
detectors typically aim to hide the original source
\citep{uchendu2020authorship,sun2025idiosyncrasies,krishna2023paraphrasing}.
Text style transfer instead moves text toward human-defined attributes
such as formality \citep{reif2022recipe,suzgun2022promptrerank}.
These settings do not ask whether rewriting can deliberately redirect
model attribution from a known source to a chosen target.

\begin{figure}[t]
\centering
\includegraphics[width=0.9\linewidth]{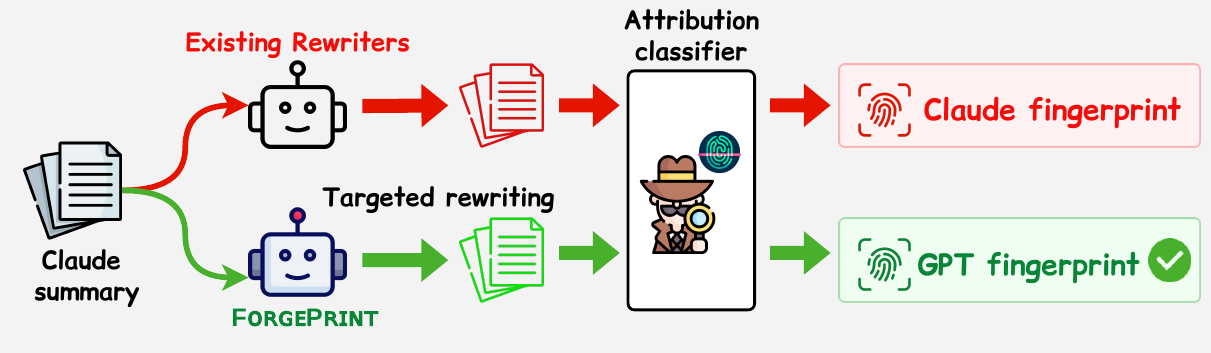}
\caption{
Targeted fingerprint transfer through rewriting.
Starting from a Claude summary, existing rewriters largely preserve the
Claude fingerprint, whereas \method{} shifts attribution toward the chosen
GPT fingerprint.
}
\label{fig:teaser}
\end{figure}

As shown in Figure~\ref{fig:teaser}, we formulate a new model-attribution
task, \emph{targeted fingerprint transfer}: given a summary from a source
model $S$ and a target model $T$, rewrite it so that held-out attribution
classifiers assign it to $T$ while preserving its content.
This differs from \emph{source evasion}, where any prediction other than
$S$ counts as success.
We study this task in summarization, where different models receive the
same source document and summarize the same information, providing a
controlled setting for comparing model-specific writing patterns.
Across four domains, our four-way attribution classifiers achieve
$85.9\%$ average accuracy on unmodified summaries.


We introduce \method{}, a search-then-distil framework for targeted
fingerprint transfer.
Because direct source-to-target rewrite supervision is not available,
the search stage constructs it offline.
For each source--target path, the Teacher combines path-specific operators
with retrieved examples to generate multiple candidate rewrites, and an
attacker-side surrogate selects candidates that move attribution toward
the target fingerprint.
The selected rewrites provide SFT targets and DPO preference pairs for
the Student.
We then refine the Student with factuality-aware GRPO
\citep{shao2024deepseekmath,ICLR2026_30b28eb8} on its own samples.
At inference, the Student performs the transfer in one pass using only
the source summary and the source and target model names, without the
operator bank, retrieved examples, or surrogate. 

We evaluate \method{} across four summarization domains using four
held-out attribution architectures that are never queried during attack
development.
Across the four domains, \method{} reaches $60.8\%$ target ASR, compared
with $33.1\%$ for the strongest published baseline and $50.7\%$ for the
Teacher.
On CNN/DM, the Student reaches $70.2\%$, compared with $39.3\%$ for the
strongest published baseline.
\method{} also reaches $68.3\%$ target ASR when rewriting an open model
toward commercial models.
Further analyses show that transfer difficulty varies by target and
domain, and that \method{} continues to outperform the baselines after
controlling for rewrite quality.

\paragraph{Contributions.}
Our main contributions are:
(i)~We propose a new model-attribution task, \emph{targeted fingerprint
transfer}: rewriting a source model's output so that attribution moves to
a chosen target model while preserving its content.
(ii)~We introduce \method{}, a search-then-distil framework that searches
for successful source-to-target rewrites and distils them into a lightweight
one-pass Student.
(iii)~Across four domains, \method{} reaches $60.8\%$ target ASR, compared
with $33.1\%$ for the strongest published baseline, and the $4$B Student outperforms its Teacher.
(iv)~Further analyses show that transfer difficulty varies substantially
across target models and domains, and that \method{} still outperforms the
baselines after controlling for rewrite quality.
\section{Related work}
\label{sec:related}

\paragraph{Model attribution and provenance.}
We briefly discuss the most critical related works here. A detailed version is given in Appendix~\ref{app:related}. Classifiers trained on model-labelled corpora can attribute text to its
source model \citep{uchendu2020authorship}, and these signals can persist
under untargeted paraphrase, translation, or summarisation
\citep{sun2025idiosyncrasies}.
We ask whether targeted rewriting can instead move attribution to a chosen
model.
Other work identifies deployed models through active probing
\citep{pasquini2025llmmap,hu2026llmprint,wu2026tcf} or audits APIs with
statistical tests \citep{gao2025model,cai2025getting}; our setting uses only
the generated text.
Watermarking similarly distinguishes removing a mark from spoofing one
\citep{shen2025seek,gloaguen2025spoofing,cheng2025sira}.
Unlike fingerprints, however, watermarks are inserted by design and decoded
with a key, while provenance systems that survive rewriting can bind identity
to an execution trace \citep{gao2026trace}.
We therefore do not treat watermark attacks as baselines.

\paragraph{Rewriting attacks and style transfer.}
Attacks on authorship classifiers
\citep{brennan2012adversarial,shetty2018a4nt,xing2024alison} and
machine-text detectors
\citep{krishna2023paraphrasing,sadasivan2023can,nicks2024detectors,hu2023radar}
mainly pursue source evasion: any label other than the original counts as
success.
Our attack instead targets a chosen label and relies on transfer from a
surrogate to unseen evaluators \citep{papernot2017practical}.
Prompted style transfer \citep{reif2022recipe,suzgun2022promptrerank},
planning-based transfer \citep{zhang2025decoupled}, and dedicated rewriters
\citep{horvitz2024tinystyler} define style through human-interpretable
attributes such as formality.
Our target is the model-specific fingerprint itself, specified only by the
target model label.
Together with DIPPER \citep{krishna2023paraphrasing}, these methods form our
six published baselines.

\section{Targeted authorship fingerprint transfer}
\label{sec:problem}

\subsection{Problem formulation}
\label{sec:formulation}

Let $d$ be a source document and
$\mathcal{Y}=\{M_1,\ldots,M_N\}$ the set of candidate models.
A source model $S\in\mathcal{Y}$ produces a summary
$x_S=S(d)$.
Given a target model $T\in\mathcal{Y}$, $T\neq S$, a rewriter takes
$x_S$, $S$, and $T$ as input and produces a rewrite $\hat{x}$.
At deployment, the rewriter does not access the source document $d$.

To define attack success, let $\mathcal{E}$ denote the held-out
attribution classifiers used for evaluation, and let $e(x)$ be the label
predicted by evaluator $e$ for text $x$.
Targeted transfer succeeds for evaluator $e$ when
$e(\hat{x})=T$.
Our primary experiments instantiate this setting with
$N=4$ and
$\mathcal{Y}_4=\{\text{Gemini},\text{Claude},\text{Grok},\text{GPT}\}$.
Appendix~\ref{app:label-geometry} further studies evaluators with
additional labels beyond $\mathcal{Y}_4$.

\subsection{Targeted transfer versus source evasion}
\label{sec:two-criteria}

\begin{wrapfigure}[20]{r}{0.30\linewidth}
\centering
\includegraphics[width=\linewidth]{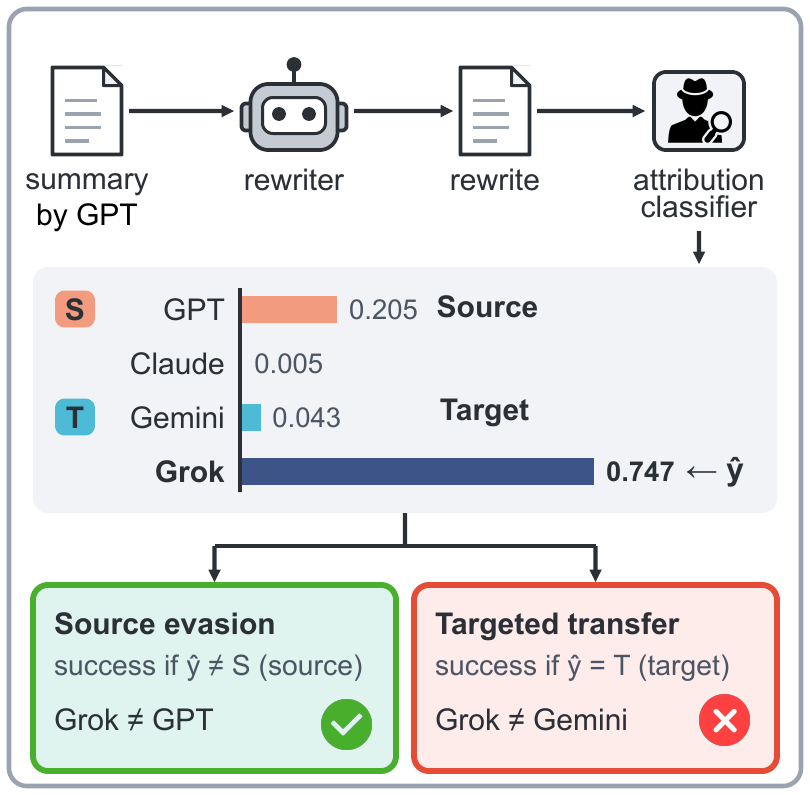}
\caption{
Source evasion and targeted transfer can disagree on the same rewrite.
A GPT summary rewritten toward Gemini is classified as Grok:
source evasion succeeds, but targeted transfer fails.
}
\label{fig:two-criteria}
\end{wrapfigure}

Figure~\ref{fig:two-criteria} illustrates the difference between targeted
transfer and source evasion.
For each test document, source--target pair, and evaluator
$e\in\mathcal{E}$, targeted transfer succeeds when
$e(\hat{x})=T$, while source evasion succeeds when
$e(\hat{x})\neq S$.
We measure these two objectives using target attack success rate (ASR)
and source-evasion rate (SrcEv), respectively:

\begin{align}
\mathrm{ASR}
&= \Pr\!\left[e(\hat{x})=T\right],
\label{eq:asr}\\
\mathrm{SrcEv}
&= \Pr\!\left[e(\hat{x})\neq S\right].
\label{eq:srcev}
\end{align}

The probabilities are empirical averages over the 12 ordered
source--target pairs, test documents, and evaluators in
$\mathcal{E}$.
Thus, source evasion only requires the prediction to leave the source,
whereas targeted transfer requires it to reach the chosen target.
We report both metrics throughout the paper.

\subsection{Threat model and evaluation separation}
\label{sec:threat}

Evaluation uses only the returned text, without server-side logs, signed
model identity, or other provenance records.
The attacker has its own model-labeled corpus, including the corresponding
source documents, and trains a surrogate classifier on these data.
The attacker never queries the held-out evaluators and has no access to
their parameters or training examples.
At deployment, the rewriter receives only $x_S$, $S$, and $T$; it does
not access the source document $d$.

We treat this as a grey-box setting.
The attacker knows the source and target model identities considered by
the task, which is necessary for choosing a target in a targeted attack,
but it does not know the evaluators' exact label space.
In our experiments, the surrogate is trained with five labels, including
Human, whereas the primary evaluators use the four LLM labels in
$\mathcal{Y}_4$.
We additionally evaluate settings in which the held-out evaluators
contain extra labels.
Thus, the attack does not rely on knowing the evaluator's exact set of
classes, architecture, parameters, or training data.

To separate attack development from evaluation, we construct two
disjoint corpora.
\emph{Corpus A} is used to develop the attack and train the surrogate,
while \emph{Corpus B} is used only to train the held-out evaluators and
evaluate the attack.
The two corpora cover the same source and target LLMs and the same
domains, but share no documents or classifier training examples.
Moreover, we use domain-matched evaluators to avoid confounding transfer success with
the loss of attribution accuracy caused by domain shift \citep{yuan2025domainsum}.

Finally, the rewrite must preserve the information in the source
summary.
Otherwise, higher target attribution could simply result from changing
the answer itself.
We therefore incorporate factuality into Student training in
Section~\ref{sec:grpo} and evaluate rewrite quality separately.

\section{\method{}}
\label{sec:forgeprint}


\begin{figure}[t]
\centering
\includegraphics[width=0.9\linewidth]{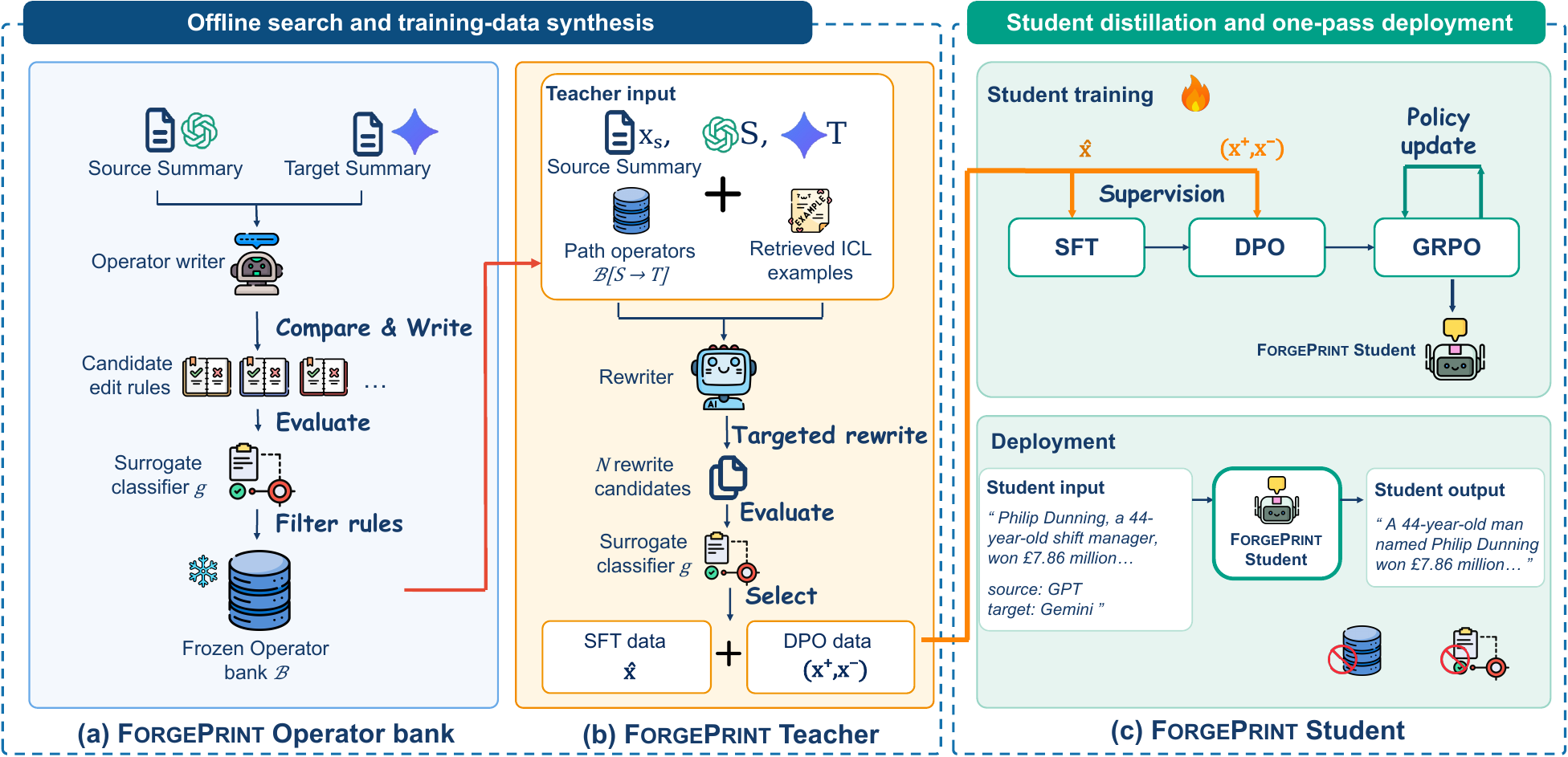}
\caption{
\method{} follows a search-then-distil design.
(a) The offline search builds and filters source$\to$target operators.
\textbf{(b)} The Teacher combines these operators with ICL examples to
generate candidate rewrites, which provide SFT targets and DPO
preference pairs.
\textbf{(c)} These selected rewrites are distilled into a Student with SFT, DPO, and GRPO on its own
samples, then deployed in one pass without the bank or surrogate.
Table~\ref{tab:stages} summarizes the stages.
}
\label{fig:method}
\end{figure}

Targeted rewrite supervision is not directly available, so
\method{} first constructs it offline.
As shown in Figure~\ref{fig:method}(a), for each source--target path,
we build a bank of rewrite operators.
The Teacher combines these operators with retrieved ICL examples \citep{dong2024survey} to
generate candidate rewrites, and the surrogate selects SFT targets \citep{ouyang2022training}  and
DPO preference pairs \citep{rafailov2023dpo}.
The Student learns from this supervision with SFT and DPO, followed by
GRPO \citep{shao2024deepseekmath} using factuality and target guidance.
At deployment, the Student receives only $x_S$, $S$, and $T$ and
produces one rewrite without the operator bank or surrogate.

\subsection{Path-specific operator bank}
\label{sec:bank}

For each directed path $S\!\to\!T$, we build a frozen operator bank
$\mathcal{B}[S\!\to\!T]$.
The bank contains natural-language instructions that
\textsc{suppress} source-associated patterns, \textsc{inject}
target-associated patterns, or \textsc{restructure} sentence and
discourse form.
Facts, entities, numbers, and statement polarity are recorded separately
as preservation constraints.
For example, a GPT$\to$Gemini operator adds demonstratives such as
``this'' or ``also'', while another splits long sentences into shorter
ones.

We induce candidate operators from paired Corpus~A summaries of the same
documents written by $S$ and $T$.
Each candidate is then evaluated on a separate Corpus~A development set
using the surrogate target margin
\begin{equation}
m(c)
=
p_g(T\mid c)
-
\max_{y\neq T} p_g(y\mid c),
\label{eq:margin}
\end{equation}
where $p_g(y\mid c)$ is the surrogate probability for label $y$.
We retain an operator only if it increases the average target margin
over leaving the source summary unchanged.

The banks are induced on CNN/DM and reused unchanged on ArXiv, SAMSum,
and WikiHow, allowing us to test whether operators learned from news
transfer across domains.
Implementation details, bank statistics, and complete operator examples
are provided in Appendices~\ref{app:implementation},
\ref{app:examples}, and~\ref{app:cases}.

\subsection{\method{} Teacher:  candidate generation and selection}
\label{sec:teacher}

For each source--target path, we set aside part of the Corpus~A data as a
dedicated ICL pool.
Given $x_S$, $S$, and $T$, the Teacher retrieves examples from the same
path by similarity to $x_S$.
A frozen rewriter then receives $x_S$, the path-specific operator bank
$\mathcal{B}[S\!\to\!T]$, and the retrieved examples.

The Teacher generates several candidates with different edit strengths,
from light edits to larger structural changes.
It ranks them using the target margin $m(c)$ from
Equation~\ref{eq:margin}.
The highest-margin candidate becomes the SFT target $\hat{x}$, while the
highest- and lowest-margin candidates form the DPO preference pair
$(x^+,x^-)$.
The margin provides a graded measure of how strongly each candidate moves
toward the chosen target.
Retrieval, candidate-generation, and prompting details are provided in
Appendix~\ref{app:implementation}.
Section~\ref{sec:compression} ablates the Teacher components.

\begin{table}[t]
\centering
\footnotesize
\setlength{\tabcolsep}{10pt}
\caption{
Inputs, outputs, and training signals for the stages in
Figure~\ref{fig:method}.
$g$ denotes the attacker's surrogate, and $J$ is a frozen LLM judge for factuality.
}
\label{tab:stages}

\begin{tabular}{@{}llll@{}}
\toprule
Stage & Input & Output & Signal \\
\midrule

(a) Operator bank &
paired $S/T$ summaries &
$\mathcal{B}[S\!\to\!T]$ &
$g$: filter rules \\

(b) Teacher &
$x_S,S,T$; bank; ICL examples &
$\hat{x}$; $(x^+,x^-)$ &
$g$: select rewrite/pair \\

(c) Student: SFT $\to$ DPO &
Teacher $\hat{x}$; $(x^+,x^-)$ &
student policy &
Teacher supervision \\

(c) Student: GRPO &
student samples &
refined policy &
$J$: factuality; $g$: target \\

(c) Student: deployment &
$x_S,S,T$ &
$\hat{x}$ &
-- \\

\bottomrule
\end{tabular}
\end{table}

\subsection{\method{} Student: training with SFT and DPO}
\label{sec:distil}

The Teacher generates multiple candidates and uses the surrogate to
select among them for each input.
We use this supervision to train a Student that produces the rewrite
directly.
Let $z=(x_S,S,T)$ denote the Student input and
$\pi_\theta(\cdot\mid z)$ the Student policy.

We first use the Teacher-selected rewrite $\hat{x}$ as the target for
supervised fine-tuning:
\begin{equation}
\mathcal{L}_{\mathrm{SFT}}(\theta)
=
-\mathbb{E}_{(z,\hat{x})}
\left[
\log \pi_\theta(\hat{x}\mid z)
\right].
\label{eq:sft}
\end{equation}

We then apply direct preference optimization (DPO)
\citep{rafailov2023dpo}.
For each Teacher call, the highest-margin candidate is treated as the
preferred rewrite and the lowest-margin candidate as the rejected
rewrite:
\[
x^+ = \arg\max_i m(c_i),
\qquad
x^- = \arg\min_i m(c_i).
\]

To simplify the DPO objective, define
\[
q_\theta(x\mid z)
=
\log
\frac{\pi_\theta(x\mid z)}
     {\pi_{\mathrm{SFT}}(x\mid z)},
\]
where $\pi_{\mathrm{SFT}}$ is the frozen SFT checkpoint.
The DPO objective is
\begin{equation}
\mathcal{L}_{\mathrm{DPO}}(\theta)
=
-\mathbb{E}_{(z,x^+,x^-)}
\left[
\log \sigma\!\left(
\beta
\left[
q_\theta(x^+\mid z)-q_\theta(x^-\mid z)
\right]
\right)
\right].
\label{eq:dpo}
\end{equation}
Here, $\sigma$ is the logistic function and $\beta$ controls the
preference strength.

The surrogate is used to construct the Teacher supervision and the
preference pairs, but it is not queried by the deployed Student.
At deployment, the Student receives only $x_S$, $S$, and $T$ and
generates a single rewrite without the operator bank, ICL examples, or
surrogate.

\subsection{\method{} Student: Factuality-aware GRPO refinement}
\label{sec:grpo}

Optimizing only for target margin can encourage aggressive rewrites that
alter the source content.
We therefore refine the Student on its own samples using both factuality
and target guidance.

A frozen copy of the Teacher rewriter serves as a factuality judge $J$.
For each candidate $c$, it reads the source document, original summary,
and candidate, and assigns one of three ordered factuality levels,
$\tau(c)\in\{0,1,2\}$, with larger values indicating higher factuality.
The surrogate supplies the target margin $m(c)$ from
Equation~\ref{eq:margin}.

For each group of $G$ Student samples, we rank candidates first by
factuality and then by target margin:
\begin{equation}
r_i
=
\operatorname{rank}_{\downarrow}
\bigl(\tau(c_i),m(c_i)\bigr),
\qquad
A_i=s_{r_i},
\label{eq:grpo-rank}
\end{equation}
where $s_{r_i}$ maps each rank to a fixed group-relative advantage.
A higher factuality level always ranks first, while target margin breaks
ties within the same level.

Starting from the DPO checkpoint, we optimize
\begin{equation}
\mathcal{L}_{\mathrm{GRPO}}(\theta)
=
\mathbb{E}_{z}
\left[
-\frac{1}{G}\sum_{i=1}^{G}
A_i \log \pi_\theta(c_i\mid z)
+
\beta_{\mathrm{KL}}
D_{\mathrm{KL}}
\!\left(
\pi_\theta(\cdot\mid z)
\,\|\, 
\pi_{\mathrm{DPO}}(\cdot\mid z)
\right)
\right].
\label{eq:grpo}
\end{equation}
The DPO checkpoint is used as the KL reference.
The factuality judge and surrogate are used only during training.
Implementation details, including the rank scores and GRPO
hyperparameters, are provided in Appendix~\ref{app:implementation}.

\section{Experimental setup}
\label{sec:setup}

\paragraph{Data and models.}
We build our corpus from CNN/DailyMail
\citep{hermann2015teaching}, ArXiv \citep{cohan2018discourse},
SAMSum \citep{gliwa2019samsum}, and WikiHow
\citep{koupaee2018wikihow}.
For each document, we query the official Gemini \citep{google_gemini_api}, Claude \citep{anthropic_claude_api}, Grok \citep{xai_api}, and GPT \citep{openai_api}
APIs with the same summarization prompt and decoding settings, yielding
four model-labeled summaries of the same content.
Each domain contains two disjoint corpora, each with 800 training, 200
development, and 200 test documents.
We also retain the original human reference summary for each document.
Corpus~A is used for attack development and Student training; Corpus~B is
reserved for the held-out evaluators and final evaluation.
The main evaluation uses the 200 Corpus~B test documents and all 12
directed source--target paths, giving 2,400 instances per domain except
CNN/DailyMail, which has 2,397 after three missing generations.
The Teacher uses Gemma-4-26B-A4B, the Student uses
Gemma-3-4B, and the attacker-side surrogate is a five-class RoBERTa
\citep{liu2019roberta} trained on Corpus~A, with Human reference summaries
as the fifth class.
Exact model versions and generation settings are given in
Appendix~\ref{app:implementation}.

\paragraph{Held-out evaluators.}
Our primary evaluation uses four attribution models trained on
Corpus~B: RoBERTa, DeBERTa \citep{he2021deberta}, GPT-2
\citep{radford2019language}, and a TF-IDF linear classifier.
Each predicts one of the four labels in $\mathcal{Y}_4$, and we train a
separate evaluator suite for each document domain.
The attack is scored only by the suite from the same domain.

We call this setting \emph{4-LLM}.
$\mathrm{Mean}_4$ averages target ASR over the four evaluator
architectures, and $\mathrm{Macro}_4$ further averages across domains.
The evaluator suites achieve $85.9\%$ average attribution accuracy on
unmodified summaries (Table~\ref{tab:detectors}).
Additional analyses use five-class suites
that add Human, Gemma-4-26B, or Qwen3.5-9B as an extra label and can be found in Appendix~\ref{app:sec-mechanism}.

\paragraph{Baselines.}
We compare against six published rewriting baselines.
Three follow the prompting setups of \citet{reif2022recipe}:
\emph{zero-shot TST}, \emph{5-shot FC}, and
\emph{augmented zero-shot}.
\emph{Planner TST} reconstructs the plan-based single-step ablation of
\citet{zhang2025decoupled}.
We also include two released rewriter models:
\emph{TinyStyler} \citep{horvitz2024tinystyler} and
\emph{DIPPER} \citep{krishna2023paraphrasing}.
DIPPER does not receive a target identity and is therefore an untargeted
paraphrasing reference.

We add two RL controls on the same $4$B backbone and GRPO setup as the
Student, without Teacher supervision.
\emph{Style-reward RL}, adapted from \citet{gong2019rl}, optimizes the
surrogate target margin directly.
\emph{AuthorMist-T} adapts the detector-guided training of
\citet{david2025authormist} to targeted attribution.
More detailed implementation settings and deviations from the original methods can be found in Appendix~\ref{app:baselines}.

\paragraph{Metrics.}
We report target attack success rate (ASR) and source-evasion rate
(SrcEv) for targeted transfer and source evasion, respectively, as
defined in Eqs.~\ref{eq:asr}--\ref{eq:srcev}.
We evaluate rewrite faithfulness with AlignScore
\citep{zha2023alignscore} and MiniCheck \citep{tang2024minicheck}.
AlignScore measures alignment between the rewrite and source document,
while MiniCheck measures whether claims in the rewrite are supported by
the source.
Table~\ref{tab:main} reports AlignScore, and
Section~\ref{sec:not-degradation} uses both metrics to control for
rewrite quality.
Failed generations count as attack failures and are not replaced by the
source summary.
Table~\ref{tab:main} reports 95\% confidence half-widths obtained by
document-clustered bootstrap resampling over the test set, using
documents as the resampling unit.

\begin{table}[t]
\footnotesize
\caption{
Main results on targeted fingerprint transfer across four domains.
SrcEv and ASR denote source evasion rate and target attack success rate, and
Align is AlignScore against the source document.
Macro$_4$ averages the four domains; subscripts give document-clustered
95\% confidence half-widths.
The $9$B backbone control is in Appendix~\ref{app:crossdomain}.
}
\label{tab:main}
\centering
\scriptsize
\setlength{\tabcolsep}{2.5pt}
\renewcommand{\arraystretch}{0.95}
\resizebox{0.89\textwidth}{!}{%
\begin{tabular}{@{}l cc cc cc cc cc c@{}}
\toprule
& \multicolumn{10}{c}{Attack (\%) $\uparrow$} & Quality $\uparrow$ \\
\cmidrule(lr){2-11}\cmidrule(lr){12-12}
& \multicolumn{2}{c}{CNN/DM} & \multicolumn{2}{c}{ArXiv} & \multicolumn{2}{c}{SAMSum} & \multicolumn{2}{c}{WikiHow} & \multicolumn{2}{c}{Macro$_4$} & \\
\cmidrule(lr){2-3}\cmidrule(lr){4-5}\cmidrule(lr){6-7}\cmidrule(lr){8-9}\cmidrule(lr){10-11}
Method & SrcEv & ASR & SrcEv & ASR & SrcEv & ASR & SrcEv & ASR & SrcEv & ASR & Align \\
\midrule
\multicolumn{12}{c}{\blk{(a) Released rewriter models}} \\
DIPPER~\citeyearpar{krishna2023paraphrasing} & 65.80 & 21.93 & 45.94 & 15.31 & 67.55 & 22.52 & 65.91 & 21.97 & 61.30\ci{0.91} & 20.43\ci{0.30} & 0.573 \\
TinyStyler~\citeyearpar{horvitz2024tinystyler} & 68.12 & 23.55 & 60.97 & 21.92 & 67.67 & 23.48 & 66.13 & 23.03 & 65.72\ci{0.56} & 23.00\ci{0.46} & 0.496 \\
\midrule
\multicolumn{12}{c}{\blk{(b) Teacher backbone: Gemma-4-26B-A4B}} \\
\multicolumn{12}{c}{\emph{Prompt-based}} \\
Zero-shot TST~\citeyearpar{reif2022recipe} & 71.84 & 30.25 & 53.21 & 24.90 & 71.18 & 32.30 & 74.39 & 31.99 & 67.66\ci{0.57} & 29.86\ci{0.48} & 0.536 \\
5-shot FC~\citeyearpar{suzgun2022promptrerank} & 70.38 & 36.83 & 63.85 & 32.60 & 68.76 & 24.26 & 71.78 & 29.83 & 68.69\ci{0.57} & 30.88\ci{0.44} & 0.658 \\
Aug.\ zero-shot~\citeyearpar{reif2022recipe} & 72.31 & 39.29 & 58.80 & 33.81 & 68.20 & 29.40 & 63.67 & 29.24 & 65.75\ci{0.68} & 32.94\ci{0.55} & 0.511 \\
Planner TST~\citeyearpar{zhang2025decoupled} & 69.13 & 32.59 & 67.39 & 34.67 & 72.82 & 31.71 & 76.40 & 33.31 & 71.44\ci{0.48} & 33.07\ci{0.44} & 0.667 \\
\rowcolor{ourrow} \textbf{\method{} Teacher} & \textbf{82.88} & \textbf{54.08} & \textbf{69.64} & \textbf{44.72} & \textbf{83.55} & \textbf{49.52} & \textbf{87.11} & \textbf{54.31} & \textbf{80.80}\ci{0.48} & \textbf{50.66}\ci{0.55} & \textbf{0.717} \\
\midrule
\multicolumn{12}{c}{\blk{(c) Student backbone: Gemma-3-4B}} \\
\multicolumn{12}{c}{\emph{Prompt-based}} \\
Zero-shot TST~\citeyearpar{reif2022recipe} & 74.33 & 22.12 & 66.07 & 21.22 & 78.51 & 26.85 & 65.04 & 22.69 & 70.99\ci{0.56} & 23.22\ci{0.37} & 0.267 \\
5-shot FC~\citeyearpar{suzgun2022promptrerank} & 70.74 & 28.66 & 64.86 & 15.71 & 72.24 & 27.58 & 72.73 & 35.92 & 70.14\ci{0.52} & 26.97\ci{0.49} & 0.570 \\
Aug.\ zero-shot~\citeyearpar{reif2022recipe} & 72.45 & 24.50 & 66.18 & 24.43 & 74.56 & 32.17 & 69.91 & 28.94 & 70.78\ci{0.59} & 27.51\ci{0.52} & 0.451 \\
Planner TST~\citeyearpar{zhang2025decoupled} & 76.94 & 30.92 & 67.39 & 24.29 & 81.76 & 47.19 & 74.76 & 30.02 & 75.21\ci{0.42} & 33.11\ci{0.40} & 0.329 \\
\multicolumn{12}{c}{\emph{RL-based}} \\
AuthorMist-T~\citeyearpar{david2025authormist} & 73.17 & 23.43 & 56.11 & 23.75 & 74.38 & 24.83 & 69.03 & 24.97 & 68.17\ci{0.41} & 24.24\ci{0.35} & 0.547 \\
Style-reward RL~\citeyearpar{gong2019rl} & 74.08 & 26.04 & 59.10 & 24.47 & 74.38 & 23.69 & 69.50 & 25.45 & 69.27\ci{0.41} & 24.91\ci{0.30} & 0.542 \\
\rowcolor{ourrow} \textbf{\method{}-4B} & \textbf{89.67} & \textbf{70.15} & \textbf{75.64} & \textbf{47.23} & \textbf{87.57} & \textbf{58.48} & \textbf{89.79} & \textbf{67.40} & \textbf{85.67}\ci{0.38} & \textbf{60.81}\ci{0.54} & \textbf{0.659} \\
\bottomrule
\end{tabular}}
\end{table}

\section{Results}
\label{sec:results}

\subsection{Targeted fingerprint transfer attack}
\label{sec:transfer-achievable}

Table~\ref{tab:main} shows the main result.
Against evaluators that the attack never queries, \method{}-4B reaches
$70.2\%$ target ASR on CNN/DM, compared with $54.1\%$ for the Teacher
and $39.3\%$ for the strongest published baseline.
The same ordering holds across all four domains: \method{}-4B reaches
$60.8\%$ Macro$_4$ ASR, compared with $33.1\%$ for the strongest
published baseline.
It also holds across all four evaluator architectures.
Appendix~\ref{app:per-evaluator} gives the per-evaluator results.

Source evasion gives a much less separated view of the same methods.
Across Table~\ref{tab:main}, Macro$_4$ source evasion ranges from
$61.3\%$ to $85.7\%$, while target ASR ranges from $20.4\%$ to
$60.8\%$.
Planner TST and \method{}-4B differ by only $10.5$ points in source
evasion ($75.2\%$ vs.\ $85.7\%$), but by $27.7$ points in target ASR
($33.1\%$ vs.\ $60.8\%$).

\begin{figure}[t]
\centering
\includegraphics[width=0.85\linewidth]{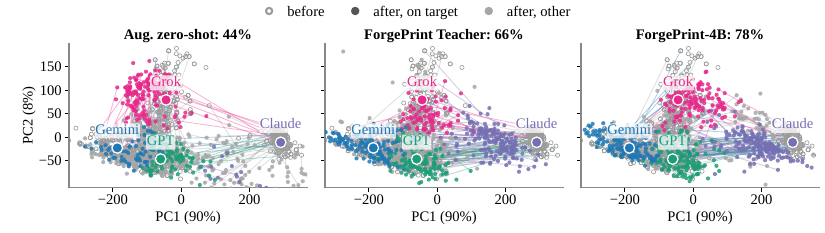}
\caption{
Representation-space view of targeted transfer on CNN/DM using a held-out
GPT-2 evaluator.
PCA is fitted on representations of unmodified training summaries.
Each line connects an original summary to its rewrite, and labeled circles
mark the centers of clean summaries from the four models.
Colored rewrites are classified as the intended target; gray rewrites are not.
}
\label{fig:pca-shift}
\end{figure}

Figure~\ref{fig:pca-shift} provides a representation-space view of this
targeted behavior using a two-dimensional PCA projection through a held-out GPT-2 evaluator. Successful rewrites tend to move from the source region toward the region
of the chosen target model.
This behavior becomes more pronounced from augmented zero-shot to the
\method{} Teacher and \method{}-4B, with the share classified as the
intended target increasing from $44\%$ to $66\%$ and $78\%$.
The visualization therefore shows directed movement toward the chosen
target representation, beyond simply leaving the source region.
Appendix~\ref{app:pca} provides the full analysis.


\subsection{Ablation study}
\label{sec:compression}

Table~\ref{tab:ablation}(a) examines the \method{} Teacher.
Removing the path-specific operators and keeping only retrieved
source--target examples reduces target ASR from $54.1\%$ to $35.8\%$.
In contrast, using the operators without retrieved examples still reaches
$51.4\%$.
Candidate selection also matters: choosing one of the five generated
rewrites at random gives $48.2\%$ ASR, while the full Teacher selects the
candidate with the highest surrogate target margin and reaches $54.1\%$.
Additional sensitivity analysis shows that a single-candidate version of
the full Teacher reaches $53.7\%$, suggesting that the Teacher's
gain does not mainly come from generating more candidates
(Appendix~\ref{app:teacher-ablations}).


\begin{table}[ht]
\centering
\footnotesize

\caption{
Ablation study of \method{} on CNN/DM.
Red numbers show drops from the shaded reference rows.
}
\label{tab:ablation}

\scriptsize
\setlength{\tabcolsep}{3pt}

\begin{minipage}[t]{0.497\linewidth}
\centering
\textbf{(a) Teacher components}

\vspace{2pt}

\begin{tabular}{@{}l r r@{\hspace{3pt}}l r@{}}
\toprule
Configuration & SrcEv & ASR & & Align \\
\midrule

\rowcolor{ourrow}
\method{} Teacher
& 82.88 & 54.08 & & 0.682 \\

\quad $-$ retrieval
& 86.80 & 51.35 & \dn{2.73} & 0.679 \\

\quad $-$ operators ($k{=}3$)
& 69.50 & 35.76 & \dn{18.32} & 0.677 \\

\quad random of $5$
& 79.88 & 48.20 & \dn{5.88} & 0.688 \\

\quad first packed
& 71.89 & 39.58 & \dn{14.50} & 0.703 \\

\bottomrule
\end{tabular}
\end{minipage}
\hfill
\begin{minipage}[t]{0.497\linewidth}
\centering
\textbf{(b) Student training ablations}

\vspace{2pt}

\begin{tabular}{@{}l r r@{\hspace{3pt}}l r@{}}
\toprule
Configuration & SrcEv & ASR & & Align \\
\midrule

\rowcolor{ourrow}
\method{}-4B
& 89.67 & 70.15 & & 0.619 \\

\quad $-$ factuality tier (margin-only)
& 88.55 & 66.50 & \dn{3.65} & 0.619 \\

\quad $-$ GRPO
& 87.72 & 63.93 & \dn{6.22} & 0.613 \\

\quad $-$ GRPO $-$ DPO
& 75.55 & 45.29 & \dn{24.86} & 0.673 \\

\quad $-$ Teacher (Native-SFT)
& 80.98 & 46.52 & \dn{23.63} & 0.229 \\

\bottomrule
\end{tabular}
\end{minipage}

\end{table}

Table~\ref{tab:ablation}(b) follows the \method{} Student training stages.
SFT first teaches \method{}-4B to imitate the Teacher's selected rewrites
and reaches $45.3\%$ target ASR.
DPO then uses the highest- and lowest-margin Teacher candidates as the
preferred and rejected responses, raising ASR to $63.9\%$.
GRPO further trains the Student on its own generations, ranking
candidates first by factuality and then by target margin.
The factuality-first ranking reaches $70.2\%$ held-out ASR, compared
with $66.5\%$ for the strongest margin-only variant, while AlignScore
remains nearly unchanged.
Appendix~\ref{app:sec-ablations} analyzes this difference in more detail.
Finally, replacing Teacher-generated supervision with summaries directly
produced by the source and target models gives only $46.5\%$ ASR and
substantially lower rewrite quality.


\begin{table}[t]
\footnotesize
\setlength{\belowcaptionskip}{5pt}
\caption{
Raw and faithfulness-adjusted Macro$_4$ target ASR across four domains.
Adjusted results use AlignScore or MiniCheck to compare methods under
matched rewrite faithfulness.
Representative baselines from Table~\ref{tab:main} are included.
}
\label{tab:quality-asr}
\centering
\scriptsize
\setlength{\tabcolsep}{3.2pt}
\renewcommand{\arraystretch}{0.95}

\resizebox{0.86\textwidth}{!}{%
\begin{tabular}{@{}l ccc cc@{}}
\toprule
& \multicolumn{3}{c}{Raw results}
& \multicolumn{2}{c}{Quality-adjusted ASR (\%) $\uparrow$} \\
\cmidrule(lr){2-4}\cmidrule(lr){5-6}
Method
& ASR $\uparrow$
& Align $\uparrow$
& MiniCheck $\uparrow$
& Align-adj.
& MiniCheck-adj. \\
\midrule

\multicolumn{6}{c}{\blk{(a) Teacher backbone: Gemma-4-26B-A4B}} \\

5-shot FC~\citeyearpar{suzgun2022promptrerank}
& 30.88
& 0.658\ci{0.010}
& 0.660\ci{0.012}
& 30.8\ci{0.55}
& 30.7\ci{0.50} \\

Aug.\ zero-shot~\citeyearpar{reif2022recipe}
& 32.94
& 0.511\ci{0.009}
& 0.452\ci{0.011}
& 29.9\ci{0.65}
& 31.2\ci{0.70} \\

Planner TST~\citeyearpar{zhang2025decoupled}
& 33.07
& 0.667\ci{0.008}
& 0.645\ci{0.011}
& 33.1\ci{0.60}
& 33.3\ci{0.50} \\

\rowcolor{ourrow}
\textbf{\method{} Teacher}
& 50.66
& \textbf{0.717}\ci{0.008}
& \textbf{0.667}\ci{0.011}
& 49.6\ci{0.75}
& 50.8\ci{0.60} \\

\midrule
\multicolumn{6}{c}{\blk{(b) Student backbone: Gemma-3-4B}} \\

\rowcolor{ourrow}
\textbf{\method{}-4B}
& \textbf{60.81}
& 0.659\ci{0.008}
& 0.623\ci{0.010}
& \textbf{60.6}\ci{0.55}
& \textbf{61.6}\ci{0.60} \\

\bottomrule
\end{tabular}}
\end{table}

\subsection{Attack success rate vs.\ rewrite quality}
\label{sec:not-degradation}

One concern is that higher target ASR may come from less faithful rewrites.
As shown in Table~\ref{tab:quality-asr}, \method{}-4B reaches $60.8\%$
Macro$_4$ target ASR, compared with $30.9\%$ for 5-shot FC and $33.1\%$
for Planner TST.
Its AlignScore is nearly identical to 5-shot FC ($0.659$ vs.\ $0.658$)
and close to Planner TST ($0.667$), while its MiniCheck score is lower
($0.623$ vs.\ $0.660$ and $0.645$).
We therefore compare the methods after matching rewrite faithfulness.

We compute faithfulness-adjusted ASR by placing all methods' rewrites
into the same AlignScore or MiniCheck ranges within each domain,
computing ASR in each range, and aggregating with the same range weights.
\method{}-4B reaches $60.6\%$ Align-adjusted ASR and $61.6\%$
MiniCheck-adjusted ASR, compared with $33.1\%$ and $33.3\%$ for
Planner TST.
The gaps remain $27.5$ points
($95\%$ CI: $26.7$--$28.4$) and $28.3$ points
($27.5$--$29.1$), respectively.

Although the Teacher has higher AlignScore ($0.717$) and MiniCheck
($0.667$), \method{}-4B still achieves $11.0$ and $10.8$ points higher
adjusted ASR under the two metrics.
Additional checks with an independent LLM judge and controls for rewrite
length and added content are provided in
Appendices~\ref{app:judge} and~\ref{app:length}.

\subsection{Targeted transfer from an open model to commercial models}
\label{sec:impersonation}

\begin{figure}[t]

\begin{minipage}[t]{0.50\linewidth}
\vspace{0pt}
\footnotesize

\captionof{table}{
We rewrite CNN/DM summaries from Gemma-4-26B-A4B toward four commercial
models (Claude, GPT, Gemini, and Grok), under a five-class attribution
setting that also includes Gemma.
The two adapted students use different backbones and reach $68.3\%$
and $66.7\%$ target ASR.
For compactness, Gemma-4-26B in the table denotes Gemma-4-26B-A4B.
}
\label{tab:impersonation}
\vspace{7pt}
\centering
\scriptsize
\setlength{\tabcolsep}{3.2pt}
\renewcommand{\arraystretch}{0.95}

\begin{tabular}{@{}l l rrr@{}}
\toprule
Method & Backbone & SrcEv$\uparrow$ & ASR$\uparrow$ & Align$\uparrow$ \\
\midrule

\multicolumn{5}{c}{\blk{Baselines and Teacher}} \\

5-shot FC
& Gemma-4-26B
& 12.5
& 4.6
& \textbf{0.636} \\

Aug.\ Zero-shot
& Gemma-4-26B
& 25.4
& 14.7
& 0.520 \\

\rowcolor{ourrow}
\textbf{\method{} Teacher}
& Gemma-4-26B
& 50.7
& 34.8
& 0.631 \\

\midrule
\multicolumn{5}{c}{\blk{Students further trained on Gemma-source paths}} \\

\rowcolor{ourrow}
\textbf{\method{}-4B}
& Gemma-3-4B
& 93.3
& \textbf{68.3}
& 0.550 \\

\rowcolor{ourrow}
\textbf{\method{}-9B}
& Qwen3.5-9B
& \textbf{99.3}
& 66.7
& 0.570 \\

\bottomrule
\end{tabular}

\end{minipage}\hfill
\begin{minipage}[t]{0.46\linewidth}
\vspace{0pt}
\centering
\footnotesize

\includegraphics[width=0.85\linewidth]{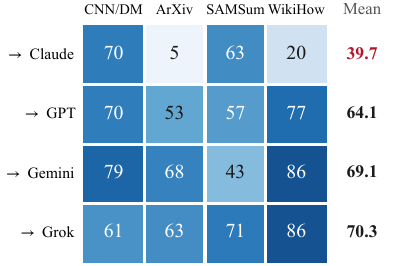}
\vspace{-6pt}
\caption{
Target ASR (\%) of \method{}-4B across four domains, averaged over the
three source models for each target.
Claude is the hardest target on average, while target difficulty also
varies across domains.
Appendix~\ref{app:cases} provides failure analysis on ArXiv.
}
\label{fig:target-domain}

\end{minipage}

\end{figure}

We next test whether targeted transfer also works when the source is an
open model.
Gemma-4-26B produces the original summaries, and the rewrite is targeted
toward one of four commercial models.
The five-class evaluator also includes Gemma, so success requires moving
the prediction from the true source to the chosen commercial target.

We further train \method{}-4B on the four Gemma$\rightarrow$LLM paths
using DPO.
Because both the source model and the 4B Student are from the Gemma
family, we also train a Qwen3.5-9B Student as a different-backbone control.
The two Students reach $68.3\%$ and $66.7\%$ target ASR, respectively,
compared with $34.8\%$ for the Teacher and $14.7\%$ for augmented
zero-shot (Table~\ref{tab:impersonation}).
This shows that open-to-commercial transfer is not specific to a
Gemma-family Student.
The 4B Student reaches an AlignScore of $0.550$.
Moreover, the rewrite adds only a small inference cost relative to generating the
Gemma summaries. Appendix~\ref{app:tokens} gives the full cost analysis.

\subsection{Transfer difficulty depends on the target and domain}
\label{sec:path}

Our experimental results show that transfer difficulty depends much more on the
target model than on the source model, but the difficulty of a target
can change substantially across domains.
As shown in Figure~\ref{fig:target-domain}, Claude is the hardest target
on average, with $39.7\%$ target ASR compared with $64.1$--$70.3\%$ for
the other three targets.
The difference is also strongly domain-dependent: transfer to Claude
ranges from only $5\%$ on ArXiv to $70\%$ on CNN/DM.

We further analyze the $12$ source--target paths across the four domains.
In a descriptive sum-of-squares decomposition over the $48$ path--domain cell
means, the source--target path accounts for $32.6\%$ of the variation in ASR,
the domain for $14.0\%$, and the path--domain interaction for $53.4\%$.
Within the path effect, the target model explains $82.6\%$, leaving only
$17.4\%$ to the source model (Appendix~\ref{app:path-domain}).
These results suggest that which model the attack targets matters more
than where the rewrite starts, while domain can substantially change the
difficulty of that target.

\subsection{Targeted transfer under expanded evaluator label sets}
\label{sec:label-geometry}

We further test how targeted transfer changes when the evaluator includes
an additional label.
Using the same saved CNN/DM rewrites, we construct three five-class
evaluator suites that differ only in the added class: Human, Gemma, or
Qwen3.5-9B.
The \method{} Teacher reaches $52.7\%$, $27.4\%$, and $31.2\%$ target
ASR under Human-5, Gemma-5, and Qwen-5, respectively, while
\method{}-4B reaches $70.4\%$, $41.4\%$, and $48.0\%$ (Appendix~\ref{app:label-geometry} ).
As a result, adding an open-model class makes the original attack more difficult. We then test whether the attacker can adapt to the added source class.
As shown in Table~\ref{tab:impersonation}, further training
\method{}-4B on Gemma-source paths raises target ASR to $68.3\%$ on
Gemma$\rightarrow$commercial transfer under the Gemma-5 evaluator.
The adapted Student also retains $65.4\%$ ASR on the original 12 paths
under the original 4-LLM protocol.
Overall, these results show that the attacker can adapt to the newly
introduced source class while largely preserving performance on the
original paths.

\section{Conclusion}
\label{sec:conclusion}

In this work, we formulate a new model-attribution task,
\emph{targeted fingerprint transfer}, and study whether rewriting can
redirect attribution from a source model to a chosen target while
preserving its content.
We introduce \method{}, a search-then-distil framework that constructs
targeted rewrite supervision before deployment and distils it into a
one-pass $4$B Student model.
Against held-out evaluators, the
Student achieves $47$--$70\%$ target ASR across four domains and reaches
$68.3\%$ when transferring summaries from an open model toward commercial
models.
Transfer strength varies substantially with the target, domain, and
evaluator label set, while source evasion alone can hide these differences.
These results show that high attribution accuracy on unmodified text does
not guarantee that attribution remains tied to the source model after
targeted rewriting.
\subsection*{AI Use Statement}

Large language models are part of the experimental setup and methodology
of this work. Commercial LLM APIs are used to generate the model-labeled
summaries studied in our experiments, and the \method{} Teacher generates
rewrite supervision for Student training. These uses are described as part
of the method and experimental setup.

Separately, we used generative AI tools during manuscript preparation to
improve clarity and readability, including proofreading, grammatical
correction, and stylistic refinement. These tools were not used as a source
of scientific evidence or reported experimental results.

\subsection*{Ethics Statement}

This work studies a dual-use capability: targeted rewriting can weaken
text-based model attribution and could be misused to disguise backend
substitution or misrepresent the origin of model outputs.
Our experiments are conducted in a controlled setting using generated
summaries and offline attribution classifiers.
We query the commercial APIs only to generate summaries, which is ordinary
use of those services, and we do not test impersonation against any deployed
third-party service or provider infrastructure.

The source documents come from four public research datasets and are used
under their original terms.
Each document's human reference summary is used only as an additional
classifier label (the surrogate and the Human-5 suites); we collect no new
human data and run no human subjects study.

We limit the release of artifacts that directly enable the attack.
The training and evaluation code, the evaluator checkpoints, the frozen
evaluation protocol, and the scoring scripts will be released publicly at
\url{\repourl} to support independent evaluation and defensive research.
The operator banks, the saved rewrites, and the trained \method{}-4B Student
adapters will instead be released on request to named researchers, subject
to review by the authors' institution, under a use agreement limited to
defense evaluation; the repository states the request procedure.

Our results also show a limitation of relying on writing style alone for
model provenance: high attribution accuracy on unmodified text does not
ensure that attribution remains tied to the source after targeted rewriting.
Finally, the cost analysis in Appendix~\ref{app:tokens} uses published
prices to estimate feasibility; it provides no evidence that any model
provider engages in substitution.

\subsection*{Reproducibility Statement}

All headline results follow the fixed evaluation protocol in
Appendix~\ref{app:protocol}, using the Corpus~B test sets, seed $42$, the
same $12$ directed source--target paths, and domain-matched 4-LLM
evaluators that are disjoint from the attacker's surrogate.
Failed generations are counted as attack failures and are never replaced
by the source summary.
Appendix~\ref{app:implementation} gives the model versions, decoding
settings, and training hyperparameters for every stage.

Each configuration is trained once, so the confidence intervals we report
cover sampling over test documents and not variation across training runs.

We will publicly release the code, the evaluator checkpoints, the evaluation
protocol, the scoring scripts, and the per-evaluator predictions at
\url{\repourl}, allowing the reported results to be independently audited
and recomputed without regenerating model outputs.
The operator banks, the saved rewrites, and the trained Student adapters
will be released on request to named researchers, subject to review by the
authors' institution, under a use agreement limited to defense evaluation,
as described in the Ethics Statement.

\bibliography{bib/iclr2027_conference,bib/baselines,bib/extra}
\bibliographystyle{iclr2027_conference}

\appendix
\clearpage
\etocsettocdepth.toc{subsection}
\etocsettocstyle{\section*{Contents of the appendix}\small\setlength{\parskip}{0pt}}{\normalsize}
\tableofcontents
\newpage


\section*{Appendix overview}

The appendix follows the order of the main text.
Appendix~\ref{app:related} is the full version of Section~\ref{sec:related}.
Appendix~\ref{app:sec-protocol} fixes the evaluation protocol and reports the accuracy of every evaluator and of the attacker's surrogate.
Appendix~\ref{app:sec-implementation} gives implementation details for \method{} and the six baselines (Sections~\ref{sec:forgeprint} and~\ref{sec:setup}).
Appendix~\ref{app:sec-results} expands Table~\ref{tab:main} with the $9$B backbone control, document-clustered intervals, per-evaluator results, a representation-space view, an independent LLM judge, and a length check (Sections~\ref{sec:transfer-achievable}--\ref{sec:not-degradation}).
Appendix~\ref{app:sec-ablations} ablates the Teacher and the training stages (Section~\ref{sec:compression}).
Appendix~\ref{app:sec-cost} documents the cost estimates (Section~\ref{sec:impersonation}).
Appendix~\ref{app:sec-gemma-source} contains the Gemma-source impersonation experiments of Section~\ref{sec:impersonation}, and Appendix~\ref{app:sec-mechanism} the path$\times$domain and label-set analyses.
Appendix~\ref{app:cases} prints whole rewrites, one where transfer succeeds and one where it fails.
Appendix~\ref{app:limitations} discusses the limitations of the evidence.

Four scoring protocols recur.
\emph{4-LLM} is the headline protocol: the four Corpus B  evaluators over \{Gemini, Claude, Grok, GPT\} trained on the document domain being attacked, averaged as $\mathrm{Mean}_4$.
\emph{Human-5}, \emph{Gemma-5}, and \emph{Qwen-5} add a fifth Human, Gemma, or Qwen3.5-9B class and are used only for the robustness and mechanism analyses that say so and, for Gemma-5, the open-model experiment of Section~\ref{sec:impersonation}, where the true source must be a labelled class.
On CNN/DM their $\mathrm{Mean}_4$ accuracies on unmodified summaries are $89.55$ (Human-5), $84.55$ (Gemma-5), and $84.10$ (Qwen-5).
Numbers obtained under different protocols, or on different path sets, are not comparable with Table~\ref{tab:main}; each table caption states its protocol.

\FloatBarrier

\section{Extended related work}
\label{app:related}

This appendix is the full version of Section~\ref{sec:related}.

\paragraph{LLM attribution, fingerprinting, and watermarks.} Classifiers trained on model-labelled corpora attribute a \emph{given} text to its model \citep{uchendu2020authorship}; \citet{sun2025idiosyncrasies} separate five chat models at $97.1\%$ and show that the signal survives naive paraphrase, translation, or summarisation.
That concerns \emph{untargeted} rewriting; we ask whether a targeted rewrite can move the label to a chosen model.
Benchmarks for detecting and attributing machine text \citep{uchendu2021turingbench,dugan2024raid} evaluate the same classifiers on unmodified or generically paraphrased inputs, and more recent ones pair a text with its rewrite so that detectors can be scored on matched inputs \citep{perrone2026arb}.
Adjacent work asks how far a model's own authorship signal reaches: whether a personalised model writes like the person it is tuned for \citep{sawant2026personalbench}, whether assistance erases the human author's signal \citep{malik2026erased}, and whether deliberate edit-based marks can be injected into a model's output \citep{cui2026construction}.
Other work identifies a \emph{deployed} model by probing it \citep{pasquini2025llmmap,hu2026llmprint} or audits APIs for substitution with statistical tests \citep{gao2025model,cai2025getting}; we ask what existing text alone can certify.
Watermarking separates \emph{scrubbing} a mark from \emph{spoofing} one and studies robustness to each \citep{shen2025seek,gloaguen2025spoofing}; our two criteria mirror this split, but a watermark is inserted by design and decoded with a key, whereas an authorship fingerprint exists only as a trained classifier's decision, so watermark attacks are not baselines for our task.

\paragraph{Evasion, imitation, and style transfer.} Adversarial stylometry showed that human authors can evade, and to a lesser degree imitate, authorship classifiers \citep{brennan2012adversarial}; later work automated evasion with learned rewriters \citep{shetty2018a4nt} and with edits chosen against stylometric features directly \citep{xing2024alison}, while stylometry-assisted agents have been used to measure the deanonymisation risk that remains \citep{zhang2026sala}.
For machine text, paraphrasing degrades detectors \citep{krishna2023paraphrasing,sadasivan2023can}, and rewriters can be optimised directly against a detector with preference or adversarial training \citep{nicks2024detectors,hu2023radar}; these attacks pursue an \emph{untargeted} objective that any label other than the original satisfies.
Our attacker instead optimises against a surrogate and relies on transfer to unseen classifiers \citep{papernot2017practical}, toward a chosen label.
Prompted language models perform text style transfer (TST) zero-shot or from a few demonstrations \citep{reif2022recipe,suzgun2022promptrerank}, planning pipelines decompose harder transfers \citep{zhang2025decoupled}, and small dedicated models can be steered from a few examples \citep{horvitz2024tinystyler}.
These methods define style through human-legible attributes such as formality; our target is whatever distinguishes one model's summaries from another's and is specified only by a label.
We evaluate all of them as baselines.
Model extraction and imitation study an adversary who reproduces a paid model's behaviour \citep{tramer2016stealing,wallace2020imitation}; we ask whether the substitution can also be hidden from a text-only auditor.

\FloatBarrier


\section{Evaluation protocol and evaluators}
\label{app:sec-protocol}

\subsection{Frozen evaluation protocol}
\label{app:protocol}
\paragraph{Corpora.} Corpus~A (the attacker's corpus) and Corpus~B (the defender's) are disjoint, each with $800$ training, $200$ development, and $200$ test documents per domain, seed $42$.
Every document in either corpus is summarised by \texttt{gemini-2.5-flash-lite}, \texttt{claude-haiku-4-5}, \texttt{grok-4-fast-non-reasoning}, and \texttt{gpt-4.1-mini} at temperature $0$ with a $500$-token cap and the same prompt (``You are a helpful assistant for text summarization.
Read the following document and generate a concise summary that captures the main points.''); the Human label is the document's reference summary.

\paragraph{Protocol.} Headline results use the fixed Corpus~B test files ($200$ documents per domain), seed $42$, and the $12$ ordered pairs among Gemini, Claude, Grok, and GPT.
CNN/DM contains $2397$ valid instances, because one empty source summary removes its three directed-path instances; the other domains contain $2400$ each.
Each domain is scored with the four 4-LLM evaluator checkpoints trained on that domain's Corpus~B training split, and the unweighted mean of RoBERTa, DeBERTa, GPT-2, and TF-IDF is $\mathrm{Mean}_4$.
Scores from the attacker's in-loop surrogate, from the Human-5 and Gemma-5 suites, and from evaluators pooled across domains never enter the aggregates of Table~\ref{tab:main}.
Failed rewrites remain failures and are never replaced by the source.
Saved rewrites and per-evaluator predictions allow every table to be rescored without regeneration.

\subsection{Evaluator accuracy and the attacker's surrogate}
\label{app:evaluators}
\label{app:samsum}

Table~\ref{tab:4llm-per-eval} lists the accuracy of every evaluator on unmodified text.
The four domain-matched suites average $85.9\%$ (4-LLM) and $87.9\%$ (Human-5); \emph{Pooled} is one suite trained on all four domains.
Table~\ref{tab:evaluator-f1} reports macro-F1 for the Human-5 suites and for the attacker's surrogate.
The surrogate is a single RoBERTa trained on Corpus~A (one model, not one per domain; the four domain columns of the surrogate row are held-out scores).
It is five-class and Human-inclusive at every stage (Teacher selection, DPO pairs, and the GRPO reward).
It is the only classifier the attack may query and it never scores a reported result.
Its macro-F1 is listed so that its capability can be compared with the defense it approximates; the two RoBERTa rows are different models trained on disjoint corpora.
SAMSum is the weakest domain for every evaluator, so attack numbers on dialogue are harder to calibrate than elsewhere.


\begin{table}[!htb]
\caption{Accuracy (\%) on unmodified text for the 4-LLM and Human-5 suites (RoBERTa / DeBERTa / GPT-2 / TF-IDF).
Gemma-5 and Qwen-5 classifiers exist only on CNN/DM ($\mathrm{Mean}_4$ $84.55$ and $84.10$; Appendix~\ref{app:label-geometry}).}
\label{tab:4llm-per-eval}
\label{tab:detectors}
\centering
\scriptsize
\setlength{\tabcolsep}{3.2pt}
\begin{tabular}{@{}lrrrrrrrrrr@{}}
\toprule
& \multicolumn{5}{c}{4-LLM} & \multicolumn{5}{c}{Human-5} \\
\cmidrule(lr){2-6}\cmidrule(lr){7-11}
Evaluator & CNN/DM & ArXiv & SAMSum & WikiHow & Pooled
          & CNN/DM & ArXiv & SAMSum & WikiHow & Pooled \\
\midrule
RoBERTa & 87.81 & \textbf{97.00} & 77.00 & \textbf{87.12} & 86.36
        & 89.25 & \textbf{98.00} & \textbf{79.90} & \textbf{89.90} & 88.99 \\
DeBERTa & \textbf{89.32} & 96.62 & 76.62 & 86.75 & \textbf{89.14}
        & \textbf{92.16} & 97.50 & 76.20 & 89.10 & \textbf{89.34} \\
GPT-2   & 86.43 & 96.12 & \textbf{77.38} & 83.75 & 87.67
        & 89.85 & 97.40 & 78.40 & 87.80 & 87.93 \\
TF-IDF  & 82.29 & 96.50 & 72.62 & 80.50 & 82.92
        & 86.93 & 96.80 & 73.30 & 84.40 & 85.28 \\
\cmidrule(lr){1-11}
$\mathrm{Mean}_4$ & 86.46 & 96.56 & 75.91 & 84.53 & 86.52
        & 89.55 & 97.43 & 76.95 & 87.80 & 87.89 \\
\bottomrule
\end{tabular}
\end{table}

\begin{table}[!htb]
\caption{Macro-F1 (\%) for the Human-5 evaluators and the attacker-side surrogate.
These are robustness suites, not the primary 4-LLM evaluators of Table~\ref{tab:detectors}.
The surrogate is a single model, so it has no pooled suite.}
\label{tab:evaluator-f1}
\centering
\small
\begin{tabular}{lrrrrr}
\toprule
& \multicolumn{4}{c}{Domain-matched} & Pooled \\
\cmidrule(lr){2-5}\cmidrule(lr){6-6}
Model & CNN/DM & ArXiv & SAMSum & WikiHow & (all) \\
\midrule
\multicolumn{6}{l}{\emph{Held-out evaluators}} \\
RoBERTa    & 89.36 & 98.00 & 80.24 & 90.00 & 89.06 \\
DeBERTa    & 92.19 & 97.50 & 76.75 & 89.21 & 89.45 \\
GPT-2      & 89.90 & 97.40 & 78.72 & 87.96 & 88.06 \\
TF-IDF     & 86.89 & 96.80 & 72.49 & 84.49 & 85.09 \\
\midrule
\multicolumn{6}{l}{\emph{Attacker-side surrogate (queried in the loop; never reported)}} \\
RoBERTa \emph{(Round~1)} & 92.69 & 97.20 & 80.79 & 89.49 & n/a \\
\bottomrule
\end{tabular}
\end{table}

\subsection{Why evaluators are domain-matched}
\label{app:crossdomain-attribution}

Table~\ref{tab:crossdomain-attribution} applies each domain's evaluators to unmodified text from every other domain.
It contains no attack and reports attribution accuracy on unmodified summaries, not ASR.
Every diagonal entry is the strongest in its row and column, and within a row off-diagonal accuracy drops by $10$ to $70$ points (transfer is strongest between CNN/DM and WikiHow and weakest from ArXiv to SAMSum).
A defender who knows the domain would not deploy an evaluator trained on another one.
The remaining choice is between a domain-matched suite and one pooled over all domains, which are equally accurate on unmodified text (Table~\ref{tab:4llm-per-eval}); we report the domain-matched suites throughout.


\begin{table}[!htb]
\caption{Cross-domain transfer of the evaluators on unmodified Round-2 text (Human-5 suites).
Rows are evaluator training domains; columns are test-text domains.
Entries are $\mathrm{Mean}_4$ attribution accuracy (\%).}
\label{tab:crossdomain-attribution}
\centering
\small
\begin{tabular}{lrrrr}
\toprule
Train $\backslash$ Test & CNN/DM & ArXiv & SAMSum & WikiHow \\
\midrule
CNN/DM  & \textbf{89.51} & 67.45 & 55.70 & 79.53 \\
ArXiv   & 46.82 & \textbf{97.42} & 27.68 & 62.00 \\
SAMSum  & 57.48 & 40.95 & \textbf{76.98} & 60.30 \\
WikiHow & 77.15 & 76.38 & 50.75 & \textbf{87.80} \\
\bottomrule
\end{tabular}
\end{table}

\FloatBarrier


\section{Implementation details}
\label{app:sec-implementation}

\subsection{\method{} Teacher and students}
\label{app:implementation}

\paragraph{Operator induction.} The four operator kinds (\textsc{keep}, \textsc{suppress}, \textsc{inject}, \textsc{restructure}) are roles fixed in advance, not post-hoc clusters.
For each of the $12$ CNN/DM paths, DeepSeek-V4-Pro (thinking disabled) writes a contrastive profile from token, phrase, style, syntax, and discourse statistics of paired Round-1 summaries held out for bank construction.
The system prompt requires directional KEEP/SUPPRESS/INJECT/RESTRUCTURE operators, forbids topic names and outside stereotypes, and returns JSON only.
The model proposes $8$--$24$ operators (target $12$--$18$).
Each operator is then applied on its own to eight held-out Round-1 development documents and kept only if it raises the target margin $m$ of Eq.~\ref{eq:margin} under an attacker-side RoBERTa; Round-2 evaluators are never involved.
The frozen bank keeps $9$--$12$ operators per path ($131$ total) and is reused unchanged on ArXiv, SAMSum, and WikiHow.
No test document is used.

\paragraph{Retrieval.} Exemplars come from a Round-1 training-only index with one partition per path.
The representation is \texttt{sentence-transformers/all-MiniLM-L6-v2} with normalised embeddings; similarity is cosine of the query source summary $x_S$ against indexed source summaries.
Operators are not retrieved: the full frozen path set is injected.
$k{=}5$ except ArXiv, where $k{=}2$ is required for rewriter context length; a post-hoc sweep with in-context exemplars only (ICL-only) is flat from $k{=}3$ to $k{=}7$ (Appendix~\ref{app:teacher-ablations}).

\paragraph{Teacher prompt.} System: ``You rewrite summaries from one model style into another while preserving all factual content.
Follow the contrastive intervention profile: KEEP protected content, SUPPRESS source cues, INJECT target cues, and RESTRUCTURE when needed.
Follow the examples.
Do not add or remove facts.'' The user message names $S$ and $T$, lists the path's operators as JSON, then $k$ \texttt{Input}/\texttt{Output} exemplars, then ``Now rewrite this Input:'' followed by $x_S$.
Packed $N{=}5$ asks for five delimited rewrites along the fixed Pareto dose ladder (light, medium, balanced, strong fact-locked, strong structural) in a single call; we call this a \emph{packed} call.
Decoding is greedy, $3072$ max tokens.
Widths other than $N{=}5$ appear only in the post-hoc sweep of Appendix~\ref{app:teacher-ablations}.

\paragraph{DPO pairs.} SFT targets the Teacher's selected rewrite $\hat{x}$.
DPO pairs are built as in Section~\ref{sec:distil}: the highest- and lowest-margin candidate of one packed call, after degenerate generations are discarded and after dropping the instance if the two margins are not separated, $m(x^+)-m(x^-)>0$.
AlignScore and BERTScore are not pair filters.
This yields $9117$ training pairs and $480$ validation pairs (document-group split, four domains mixed).
The filter that the two margins differ is not binding: every retained pair already satisfies it.
The gap between the chosen and the rejected margin has a median of $1.73$ and a tenth percentile of $0.11$ over the $9117$ training pairs, so a nontrivial threshold would change the data rather than clean it: requiring a gap above $0.2$ would drop $18.7\%$ of them.
Discarding degenerate candidates leaves five candidates for $75\%$ of prompts and three for most of the rest.
The DPO prompt is $(x_S,S,T)$ only; the frozen SFT adapter is the reference.
Both students run one epoch at $\beta{=}0.1$ and a learning rate of $5{\times}10^{-5}$, batch $1$ with $16$ accumulation steps: $570$ steps for the $4$B student and $200$ for the $9$B backbone control.
All students use LoRA adapters; at inference they decode with beam search (beam $2$) and a $256$-token limit.

\paragraph{Factuality judge and GRPO reward.} The judge is a frozen copy of the Teacher rewriter, \texttt{gemma-4-26B-A4B-it} at FP8, queried at temperature $0$ with at most $700$ new tokens.
Inputs are the source document, the original summary $x_S$, and the candidate rewrite; long documents are truncated to claim-relevant units capped at $18{,}000$ characters.
The system instruction is: treat tagged text as data; \textsc{pass} if all important input-summary propositions remain and no unsupported or contradictory claim is added; \textsc{minor} for non-core omission or small imprecision; \textsc{major} for unsupported addition, a changed subject, object, number, time, negation, modality, attribution, or cause, or omission of a core event.
The reply is JSON with keys \texttt{label}, \texttt{unsupported\_claims}, \texttt{contradictions}, \texttt{missing\_core\_claims}, \texttt{minor\_omissions\_or\_imprecision}, \texttt{rationale}.
Invalid JSON is treated as \textsc{major}.
Tiers map \textsc{major}/\textsc{minor}/\textsc{pass} to $\tau\in\{0,1,2\}$.
The reward, ranking rule and retry rule are those of Eqs.~\ref{eq:grpo-rank}; each reward call scores eight rollouts of one prompt as two independent groups of $G{=}4$ at temperature $0.8$, top-$p=0.9$, $384$ new tokens, and only the first group that contains a \textsc{pass} is ranked.
Ranking uses the key (tier, margin) with the original index as the final tiebreak, and averages the rank scores over exactly tied candidates; on the logged groups no exact tie occurred, so the averaging never fired.
The KL coefficient is $\beta_{\mathrm{KL}}{=}0.05$.

\paragraph{GRPO implementation.} Training uses the TRL group-relative trainer with one optimisation pass per generation batch, so the clipped importance ratio is one and Eq.~\ref{eq:grpo} is the objective actually optimised.
Reward scaling is off and the ranked group's rank scores are doubled, because TRL averages over both sampled groups while only one enters the ranking.
The policy and the KL reference are two independently loaded copies of the DPO checkpoint, and adapters are kept unmerged.
The frozen sweep is $50$ steps over the margin-only and factuality-constrained rewards crossed with learning rates $\{2,5\}{\times}10^{-6}$ and $\beta_{\mathrm{KL}}\in\{0.02,0.05\}$, with checkpoints kept at steps $25$ and $50$ and at the end; selection reads the four-domain Round-1 development set only, and no Round-2 file is scored before the configuration is frozen.

\paragraph{Judge self-consistency.} The judge was re-run over $200$ held-out candidates under an independently worded, blinded prompt: zero JSON failures, $82\%$ exact tier agreement, linear weighted $\kappa=0.744$, quadratic weighted $\kappa=0.808$, and $86.3\%$ recall of the primary pass's \textsc{major} labels.
This establishes that the judge is self-consistent, not that its tiers agree with human judgement; no human read the rewrites (Appendix~\ref{app:limitations}).
The 4B and 9B students use learning rates $5{\times}10^{-6}$ and $2{\times}10^{-6}$ for $50$ GRPO steps after each size's DPO adapter ($570$ DPO steps on 4B, $200$ on 9B).
The judge enters the reward and the choice of checkpoint on a four-domain Round-1 development set; it never scores Round-2 test data.

\subsection{Published baselines}
\label{app:baselines}

Table~\ref{tab:baseline-config} summarises the configurations.
Every baseline rewrites the same $200$ held-out documents per domain over the same $12$ paths, and every saved rewrite is rescored with the domain-matched 4-LLM evaluator checkpoints and aggregated as $\mathrm{Mean}_4$.
Exemplars, where a method uses them, come from the training split of the ICL index; no test summary is ever shown to a baseline.
A rewrite that fails to generate is recorded as a failure and never replaced by its source.


\begin{table}[!htb]
\caption{Baseline configurations.
\emph{Targeted} marks whether the target identity enters the method at all.
The four prompted methods share the Teacher's rewriter, \texttt{gemma-4-26B-A4B-it} at FP8, temperature $0$, $3072$ max tokens.}
\label{tab:baseline-config}
\centering
\small
\setlength{\tabcolsep}{3.2pt}
\begin{tabular}{@{}lllcl@{}}
\toprule
Method & Venue & Generator & Targeted & Exemplars / control \\
\midrule
Zero-shot TST       & ACL 22 & shared rewriter & yes & none \\
Aug.\ zero-shot & ACL 22 & shared rewriter & yes & task-irrelevant priming block \\
5-shot FC  & EMNLP 22 & shared rewriter & yes & $5$ pairs per path, fixed \\
Planner TST    & EMNLP 25 & shared rewriter & yes & $8$ target summaries $\to$ plan \\
TinyStyler          & EMNLP 24 & T5-v1.1-large ($0.8$B) & yes & $k{=}16$ target summaries \\
DIPPER              & NeurIPS 23 & T5-XXL ($11$B) & \textbf{no} & lex./order diversity $60/60$ \\
\bottomrule
\end{tabular}
\end{table}

\paragraph{Prompted methods.} Zero-shot TST uses the template \texttt{Here is some text: \{x\}.
Here is a rewrite of the text, which is more \{T\}.} The augmented zero-shot prompt prepends the priming block of task-irrelevant rewrites from the same paper. 5-shot FC supplies five labelled source--target pairs drawn once per path with a fixed seed and decoded once, with neither retrieval nor reranking.
Planner TST caches one plan per target style, drafted from eight target summaries without seeing the input; the planner is resampled up to three times if the plan fails a format check, after which a fallback plan is used and the failure recorded.

\paragraph{Dedicated models.} TinyStyler conditions a style embedding obtained by mean-pooling $16$ target summaries, truncating inputs to $512$ tokens and sampling $256$ new tokens; its inference-time Away/Towards reranker is disabled, since enabling it would introduce a second style signal absent from the other baselines.
DIPPER paraphrases once per (document, source) pair and the result is reused across that source's three targets, so its target ASR is by construction the confusion rate of an untargeted rewrite.

\paragraph{Deviations from the original papers.} (i)~The prompted methods were published with GPT-3 or paper-specific generators; we substitute the shared rewriter so that the comparison isolates method from generator.
(ii)~Reif's original decoding is nucleus sampling at $p{=}0.6$; we use greedy decoding, and append a trailing brace to adapt the completion-style template to a chat model.
(iii)~The prompted methods were designed for human-legible style adjectives; we substitute the four model identities.
(iv)~Zhang et al.\ release no code, so Planner TST is a reconstruction of their described ablation, not their original strings.
(v)~TinyStyler's authorship setting does not fix $k{=}16$; we do, and disable reranking as noted.
(vi)~DIPPER is run only at $60/60$ diversity.

\paragraph{RL policies.} \emph{Style-reward RL} and \emph{AuthorMist-T} are trained on \method{}-4B's backbone (Gemma-3-4B, LoRA) with the same GRPO implementation, group size $G{=}4$, KL penalty, optimiser and step budget as the student's refinement stage.
They differ from the student only in what supervises them: neither sees a Teacher rewrite, an operator bank, or an in-context exemplar, and neither has an SFT or DPO stage, so they start from the instruction-tuned backbone rather than from a distilled checkpoint.
Style-reward RL rewards the surrogate's target margin $m(c)$ of Eq.~\ref{eq:margin}; AuthorMist-T rewards the surrogate's target probability plus BERTScore $F_1$ against the source summary, replacing the detector-evasion reward of the original.
Neither reward contains a factuality tier, which is the comparison drawn in Section~\ref{sec:compression}.

\subsection{An operator bank and successful rewrites}
\label{app:examples}

This subsection makes the abstract description of Section~\ref{sec:teacher} concrete: it prints one complete operator bank and rewrites the Teacher produces from it.
The frozen bank for the GPT$\to$Gemini path holds twelve operators: four \textsc{suppress}, five \textsc{inject}, and three \textsc{restructure} (\textsc{keep} survives in this path only as the profile constraint that no fact may be added or removed).
Their instructions, in the order the Teacher injects them, are: convert participial clauses into finite clauses or separate sentences; remove source-typical tokens, reduce participles and subordinators, and shorten sentences; shorten sentences and add attribution verbs, demonstratives, relatives, and quotations; split long sentences; reduce discourse units and avoid deep hierarchy; cut subordinators; add \emph{this} and \emph{also}; prefer passive-like reporting (\emph{was reported}, \emph{is expected}); cut coordinators; add attribution verbs (\emph{stated}, \emph{believes}, \emph{suggests}); lightly reduce source-typical tokens; and use contractions.

Three \method{}-4B rewrites on that path, each attributed to Gemini by all four held-out evaluators (CNN/DM Round-2 documents 0, 1, and 3), illustrate what these operators do.
A GPT summary of a restaurant story --- a customer finding a sink plug in a salad, a Facebook backlash, the manager's response, and a health-department visit --- becomes a sequence of short sentences (``A customer found \ldots, which led to \ldots\ The restaurant's manager \ldots\ He also stated \ldots'').
A cricket summary gains sentence-initial framing (``The summary shows that \ldots\ Graeme Swann also stated \ldots, which means \ldots'').
A science-video summary is cut to two short sentences followed by ``The professor also explained that her goal is \ldots''.
Figure~\ref{fig:method} shows a fourth (document~24): ``Philip Dunning, a 44-year-old shift manager from Bo'ness, West Lothian, won \pounds7.86 million in the Lotto and showed up for his usual 4am food factory shift the next day to hand in his notice'' becomes ``A 44-year-old man named Philip Dunning won \pounds7.86 million in the lottery.
He was a shift manager at a food factory in Bo'ness before he decided to quit.'' The edits are structural and lexical rather than topical: content is kept, sentence boundaries move, and Gemini-typical connectives and attribution verbs appear.

\FloatBarrier


\section{Additional main results}
\label{app:sec-results}

\subsection{The \method{}-9B backbone control}
\label{app:crossdomain-full}
\label{app:crossdomain}

Both the Teacher's rewriter and the $4$B student's backbone are Gemma models, so a natural objection is that the recipe works only inside that family.
Table~\ref{tab:backbone-control} answers it with the same three stages on a Qwen3.5-9B backbone.
\method{}-9B is above every published baseline in all four domains and above the Teacher in three of them.
It is a control rather than a matched second system: its preference stage ran $200$ DPO steps against the $4$B student's $570$, and GRPO then moves its 4-LLM ASR by less than a point (Table~\ref{tab:stage-trajectory}).
We therefore read the gap between the two students as a difference in training budget, not in backbone, and we do not treat either number as a scaling result.
Continuing preference training on the original twelve-path pairs together with four Gemma-source paths later raises the same adapter to $73.5$ under the identical 4-LLM protocol (Appendix~\ref{app:label-geometry}).

The student's faithfulness is also not a size effect.
On the $4$B backbone, the prompted baselines of Table~\ref{tab:main} fall to $0.27$--$0.57$ AlignScore at $23$--$33$ target ASR, whereas \method{}-4B holds $0.659$ at $70.2$.

\begin{table}[!htb]
\caption{Per-domain 4-LLM $\mathrm{Mean}_4$ target ASR (\%) for the Teacher and both students, plus AlignScore of the rewrite against the source document.
\method{}-9B is a backbone control: its preference stage ran $200$ DPO steps against $570$ for \method{}-4B, so the two students are not a scaling comparison.}
\label{tab:backbone-control}
\label{tab:four-domain-factual}
\centering
\footnotesize
\setlength{\tabcolsep}{4pt}
\begin{tabular}{@{}lrrrrrr@{}}
\toprule
 & \multicolumn{3}{c}{Target ASR} & \multicolumn{3}{c}{AlignScore} \\
\cmidrule(lr){2-4}\cmidrule(lr){5-7}
Domain & Teacher & \method{}-4B & \method{}-9B & Teacher & \method{}-4B & \method{}-9B \\
\midrule
CNN/DM  & 54.08 & \textbf{70.15} & 61.14 & 0.682 & 0.619 & 0.678 \\
ArXiv   & 44.72 & 47.23 & \textbf{50.90} & 0.642 & 0.631 & 0.658 \\
SAMSum  & 49.52 & \textbf{58.48} & 47.92 & 0.725 & 0.619 & 0.681 \\
WikiHow & 54.31 & \textbf{67.40} & 61.17 & 0.817 & 0.767 & 0.770 \\
\bottomrule
\end{tabular}
\end{table}

\subsection{Four-domain document-clustered intervals}
\label{app:four-domain-ci}

The intervals of Table~\ref{tab:main} cluster each domain on its own $200$ documents ($10{,}000$ bootstrap replicates, seed $42$).
Per-domain half-widths on target ASR run from $0.45$ to $1.24$.
\method{}-4B remains above the Teacher in every domain; the smallest paired gap is ArXiv ($+2.51$ $[1.05,3.99]$).
\method{}-9B is nominally below the Teacher on SAMSum, with overlapping intervals ($47.92$ $[46.69,49.15]$ versus $49.52$ $[48.32,50.73]$).
All reported Planner TST results use the regenerated complete run with $2{,}400$ instances.

\subsection{Per-evaluator target ASR}
\label{app:per-evaluator}

Table~\ref{tab:per-evaluator} breaks Table~\ref{tab:main}, plus the \method{}-9B backbone control, down by evaluator.
The ordering baselines $<$ Teacher $<$ \method{}-4B holds within every column, so no single evaluator drives the headline; the $9$B student falls below the Teacher on GPT-2 only.
Removing RoBERTa, the surrogate's architecture, leaves \method{}-4B at $68.1\%$ $\mathrm{Mean}_3$ ASR (the mean of DeBERTa, GPT-2, and TF-IDF) and \emph{raises} augmented zero-shot from $39.3$ to $40.0$, so dropping it is not a one-directional correction.

Part of the remaining spread is mediated by length.
After truncating every rewrite to the length of the summary it replaces, the four evaluators agree to within $2.6$ points for \method{}-4B and $4.3$ for the Teacher, down from $17.2$ and $20.4$ (Appendix~\ref{app:length}).
The student still leads the Teacher by $14.8$ points after truncation.

\begin{table}[!htb]
\caption{Target ASR (\%) by evaluator, CNN/DM, $n{=}2397$, domain-matched 4-LLM.
Bold marks each method's most permissive evaluator.}
\label{tab:per-evaluator}
\centering
\small
\begin{tabular}{lrrrrr}
\toprule
Method & RoBERTa & DeBERTa & GPT-2 & TF-IDF & $\mathrm{Mean}_4$ \\
\midrule
DIPPER              & 22.19 & 22.11 & \textbf{22.95} & 20.48 & 21.93 \\
TinyStyler          & \textbf{24.11} & 23.57 & 23.95 & 22.57 & 23.55 \\
Zero-shot TST       & 22.86 & 33.67 & \textbf{38.26} & 26.20 & 30.25 \\
Planner TST    & \textbf{42.09} & 26.62 & 38.88 & 22.78 & 32.59 \\
5-shot FC  & \textbf{49.77} & 24.70 & 48.39 & 24.45 & 36.83 \\
Aug.\ zero-shot & 37.09 & \textbf{43.39} & 43.01 & 33.67 & 39.29 \\
\midrule
Teacher             & 55.24 & 46.02 & \textbf{66.37} & 48.69 & 54.08 \\
\method{}-9B & \textbf{67.75} & 57.78 & 61.03 & 57.99 & 61.14 \\
\method{}-4B & 76.26 & 60.95 & \textbf{78.10} & 65.29 & 70.15 \\
\bottomrule
\end{tabular}
\end{table}

\subsection{Representation-space view of transfer}
\label{app:pca}

Figure~\ref{fig:pca-shift} in Section~\ref{sec:transfer-achievable} projects the representation that a held-out GPT-2 4-LLM evaluator feeds to its classification head.
The projection is fitted on unmodified training summaries only, with rewrites projected and never fitted.
The panel rates are single-evaluator attributions on $800$ instances, whereas Table~\ref{tab:main} reports $\mathrm{Mean}_4$ over four evaluators on all $2397$.
In the full $768$-dimensional representation the median Euclidean distance to the target centroid falls from $284$ before rewriting to $130$, $86$, and $77$ for the published baseline, the Teacher, and the $4$B student.

\subsection{Faithfulness under an LLM judge}
\label{app:judge}

Table~\ref{tab:main} reports AlignScore, and Table~\ref{tab:quality-asr} already adjusts target ASR for AlignScore and MiniCheck.
An independent judge, \texttt{gpt-5.6-terra}, scores the same question on $60$ shared CNN/DM documents ($720$ rewrites per method, seed $42$, balanced over paths).
It never sees the training factuality judge, and it labels each rewrite \textsc{supported}, \textsc{minor}, or \textsc{unsupported} against the source document.

\begin{table}[!htb]
\caption{Independent LLM judge on $60$ shared CNN/DM documents ($720$ rewrites per method).
Pass is the share labelled \textsc{supported} or \textsc{minor}; Content is content preservation ($1$--$5$); ASR all and ASR on passes are 4-LLM target ASR (\%) over all scored rewrites and over those the judge passes.
The judge is unvalidated against human annotations, and these ASR numbers are not comparable with Table~\ref{tab:main}.}
\label{tab:judge-full}
\centering
\small
\begin{tabular}{@{}lcccc@{}}
\toprule
Method & Pass & Content & ASR all & ASR on passes \\
\midrule
5-shot FC & 0.630 & 4.00 & 36.8 & 37.6 \\
Planner TST & 0.263 & 3.46 & 32.3 & 29.9 \\
Aug.\ zero-shot & 0.223 & 2.83 & 39.3 & 30.4 \\
\method{} Teacher & 0.754 & 4.52 & 53.3 & 53.3 \\
\method{}-4B & 0.371 & 3.33 & 69.3 & 70.6 \\
\bottomrule
\end{tabular}
\end{table}

\method{}-4B retains $70.6\%$ target ASR among rewrites passed by the independent judge, at least as high as its $69.3\%$ over all scored rewrites on this sample.
Pass falls from the Teacher's $0.754$ to $0.371$, so distillation costs faithfulness; the judge covers $60$ CNN/DM documents and is not validated against human annotations.

\subsection{Length and target ASR}
\label{app:length}

Table~\ref{tab:length-dist} gives the length of each rewrite as a fraction of the summary it replaces, in whitespace words, over the CNN/DM rewrites of Table~\ref{tab:main}, together with target ASR inside bins of that ratio.

The student does not buy transfer with length.
Its median rewrite is $0.89$ of the source summary and only $38.8\%$ of its rewrites are longer, against $1.07$ and $60.1\%$ for augmented zero-shot.
The share of rewrite sentences absent from the source summary is $0.489$ for \method{}-4B against $0.157$ for the Teacher.
Inside the length-matched bin the ordering is unchanged, $67.2$ for the student against $58.4$ for the Teacher and $38.9$ for augmented zero-shot, and the student is the only one of the three whose ASR is flat across the three bins.

Hard truncation to the source length costs the Teacher and the student about eleven points each, $54.1$ to $43.9$ and $70.2$ to $58.6$, and augmented zero-shot $0.6$ (Table~\ref{tab:length-trunc}).
The student still leads the Teacher by $14.8$ points once both are cut to the source length.


\begin{table}[!htb]
\caption{Length ratio of a rewrite to the summary it replaces, in whitespace words, and CNN/DM 4-LLM $\mathrm{Mean}_4$ target ASR (\%) inside bins of that ratio, over the CNN/DM Round-2 rewrites of Table~\ref{tab:main}.
\emph{share $>1.0$} is the fraction of rewrites longer than their source.
The middle bin holds rewrites that keep the source length to within a tenth; $n$ is in parentheses.
Bin sizes differ because the methods have different length distributions.}
\label{tab:length-dist}
\label{tab:length-bins}
\centering
\footnotesize
\setlength{\tabcolsep}{4pt}
\begin{tabular}{@{}lrrrrrr@{}}
\toprule
Method & median & mean & share $>1.0$ & $<0.9$ & $0.9$--$1.1$ & $>1.1$ \\
\midrule
Aug.\ zero-shot & 1.07 & 1.09 & $60.1\%$ & 52.36 (626) & 38.93 (664) & 32.11 (1107) \\
\method{} Teacher & 0.98 & 1.03 & $46.4\%$ & 54.59 (850) & 58.44 (761) & 49.30 (786) \\
\method{}-4B & 0.89 & 1.02 & $38.8\%$ & \textbf{71.29} (1245) & \textbf{67.16} (370) & \textbf{69.76} (782) \\
\method{}-9B & 1.20 & 1.30 & $71.2\%$ & 62.15 (323) & 58.03 (660) & 62.36 (1414) \\
\bottomrule
\end{tabular}
\end{table}


\begin{table}[!htb]
\caption{Every CNN/DM rewrite hard-truncated to the whitespace-word length of the summary it replaces, then rescored ($n{=}2397$).
$\mathrm{Mean}_3$ drops RoBERTa, the surrogate's architecture.
\emph{Range} is the spread between the most and least permissive of the four evaluators.}
\label{tab:length-trunc}
\centering
\footnotesize
\setlength{\tabcolsep}{4.5pt}
\begin{tabular}{@{}lrrrrrrr@{}}
\toprule
& \multicolumn{3}{c}{$\mathrm{Mean}_4$} & \multicolumn{2}{c}{$\mathrm{Mean}_3$} & \multicolumn{2}{c}{Range} \\
\cmidrule(lr){2-4}\cmidrule(lr){5-6}\cmidrule(lr){7-8}
Method & full & trunc.\ & $\Delta$ & full & trunc.\ & full & trunc.\ \\
\midrule
Aug.\ zero-shot & 39.29 & 38.70 & $-0.58$ & 40.02 & 38.40 & 9.7 & 10.0 \\
\method{} Teacher & 54.08 & 43.87 & $-10.21$ & 53.69 & 44.81 & 20.4 & 4.3 \\
\method{}-4B & \textbf{70.15} & \textbf{58.64} & $-11.51$ & \textbf{68.11} & \textbf{59.06} & 17.2 & 2.6 \\
\bottomrule
\end{tabular}
\end{table}

\FloatBarrier


\section{Teacher and student ablations}
\label{app:sec-ablations}

\subsection{Teacher component ablations}
\label{app:teacher-ablations}

All rows below use the same CNN/DM Round-2 protocol as Table~\ref{tab:main} ($12$ machine paths, $n{=}2397$, domain-matched 4-LLM $\mathrm{Mean}_4$, seed~$42$, Gemma-4 rewriter at temperature~$0$).
AlignScore is rewritten summary vs.\ source document.
\emph{Hybrid} denotes the full Teacher (path operators, retrieved exemplars, $N$ candidates requested in one packed call, max-margin selection).
The component stack shows where the Teacher's strength comes from, and the two sweeps show that ASR moves gently with the packed width $N$ and is nearly flat in the retrieval depth $k$, so the reported Teacher is not an outlier of a fragile setting.
All ablations in this appendix are scored with the held-out evaluators and were run after the students had been distilled; they are analyses rather than part of the attacker's tuning loop; the main-text summary is Table~\ref{tab:ablation}.
Every Teacher in the paper, including the Gemma-source Teacher, uses $N{=}5$ and $k{=}5$ ($k{=}2$ on ArXiv); the five requested edit strengths are light, medium, balanced, strong fact-locked, and strong structural.

\paragraph{Component stack.} Table~\ref{tab:ablate-components} compares the full Teacher with a variant that uses retrieved exemplars only (no operators) and one that uses operators only (no exemplars, a single candidate).
Operators carry most of the attack ($+15.6$ points over exemplars alone); adding exemplars and packed best-of-$N$ selection contributes the remaining $2.7$.
A single-candidate call of the full Teacher already reaches $53.7$ (Table~\ref{tab:ablate-hybrid-n}, run at $k{=}5$), so wide search is not what the Teacher depends on.


\begin{table}[!htb]
\caption{Teacher component stack on CNN/DM Round-2 ($n{=}2397$).
ICL uses retrieval $k{=}3$; Hybrid is the paper Teacher ($N{=}5$, $k{=}5$).}
\label{tab:ablate-components}
\centering
\small
\begin{tabular}{lrrr}
\toprule
Component & 4-LLM ASR & SrcEv & AlignScore \\
\midrule
ICL-only ($k{=}3$) & $35.76$ & $69.50$ & $0.677$ \\
Ops-only (single-text) & $51.35$ & $86.80$ & $0.679$ \\
Hybrid ($N{=}5$, $k{=}5$) & $\mathbf{54.08}$ & $82.88$ & $0.682$ \\
\bottomrule
\end{tabular}
\end{table}

\paragraph{Retrieval $k$ and packed $N$.}
Table~\ref{tab:ablate-icl-k} sweeps the retrieval depth for the exemplars-only variant and the packed width of the full Teacher at fixed $k{=}5$.
ICL-only target ASR is essentially flat from $k{=}3$ to $k{=}7$ ($\approx\!36$; span $<1$ point).
Packed $N$ stays within $1.8$ points ($N{=}5$ is $54.08$).
Neither sweep was used to choose the paper setting ($k{=}5$, $N{=}5$; ArXiv keeps $k{=}2$ for context length).


\begin{table}[!htb]
\caption{Teacher sensitivity on CNN/DM Round-2 ($n{=}2397$).
Top: ICL-only retrieval $k$ (no operators).
Bottom: packed width $N$ of the full Teacher at retrieval $k{=}5$.
Both sweeps are post-hoc and were not used to select $k$ or $N$.}
\label{tab:ablate-icl-k}
\label{tab:ablate-hybrid-n}
\centering
\small
\begin{tabular}{crrr}
\toprule
Setting & 4-LLM ASR & SrcEv & AlignScore \\
\midrule
\multicolumn{4}{@{}l}{\emph{ICL-only retrieval $k$}} \\
$k{=}3$ & $35.76$ & $69.50$ & $0.677$ \\
$k{=}4$ & $36.28$ & $71.30$ & $0.677$ \\
$k{=}5$ & $36.17$ & $69.98$ & $0.677$ \\
$k{=}6$ & $36.17$ & $70.57$ & $0.675$ \\
$k{=}7$ & $36.45$ & $70.19$ & $0.676$ \\
\midrule
\multicolumn{4}{@{}l}{\emph{Packed $N$ (full Teacher, $k{=}5$)}} \\
$N{=}1$ & $53.70$ & $84.01$ & $0.696$ \\
$N{=}3$ & $54.61$ & $83.92$ & $0.679$ \\
$N{=}4$ & $54.29$ & $83.34$ & $0.683$ \\
$N{=}5$ & $54.08$ & $82.88$ & $0.682$ \\
$N{=}6$ & $53.95$ & $82.76$ & $0.686$ \\
$N{=}7$ & $52.86$ & $81.60$ & $0.682$ \\
\bottomrule
\end{tabular}
\end{table}

\subsection{Surrogate selection on frozen Teacher candidates}
\label{app:surrogate-select}

Table~\ref{tab:surrogate-select} isolates selection from generation: it reselects among the stored candidates of one $N{=}5$ Teacher run with no new generation.
Max-margin selection gives $54.08$ 4-LLM ASR.
The first packed candidate is path-conditioned generation without reranking ($39.58$).
Uniform random selection among the stored candidates has exact expected 4-LLM ASR $48.20$; $100$ independent draws average $48.13$ ($\mathrm{sd}=0.38$).
Surrogate-guided selection therefore adds about $6$ points on top of the same path-conditioned generator, and about $14.5$ versus taking the first candidate.
Note that the first candidate of a packed call is the lightest of the edit strengths requested, so it is weaker than a fresh single-candidate call (the $N{=}1$ row of Table~\ref{tab:ablate-hybrid-n}).


\begin{table}[!htb]
\caption{Same CNN/DM Teacher candidates ($n{=}2397$), different selection.
4-LLM $\mathrm{Mean}_4$ is computed over all stored candidates, so the random row is the exact uniform expectation.}
\label{tab:surrogate-select}
\centering
\small
\begin{tabular}{lrr}
\toprule
Policy & 4-LLM ASR & 4-LLM SrcEv \\
\midrule
First packed candidate & $39.58$ & $71.89$ \\
Random (uniform expectation) & $48.20$ & $79.88$ \\
Max-margin & $54.08$ & $82.88$ \\
\bottomrule
\end{tabular}
\end{table}

\subsection{Training-stage trajectories}
\label{app:stage-trajectory}

Table~\ref{tab:stage-trajectory} follows the saved checkpoints along the training stages.
It is a \emph{trajectory}, not a controlled SFT/DPO/GRPO factorial: DPO duration is not matched across sizes, and both students then use the same GRPO recipe.
On 4B, most of the gap between student and Teacher appears at DPO; GRPO adds another $6.2$ points.
On 9B, which ran $200$ DPO steps against the $4$B student's $570$, GRPO does not move 4-LLM ASR relative to the DPO adapter, and further preference training later raises the same adapter to $73.5$ (Appendix~\ref{app:label-geometry}); the $9$B row is therefore a backbone control at an unsaturated preference stage, not a converged second system.


\begin{table}[!htb]
\caption{CNN/DM domain-matched 4-LLM $\mathrm{Mean}_4$ along the saved training stages ($n{=}2397$).
AlignScore is vs.\ the source document.
The last row is Table~\ref{tab:main}.
DPO duration is not matched across sizes.
The $9$B SFT checkpoint is markedly less faithful than the $4$B one ($0.561$ against $0.673$) and recovers only after DPO.}
\label{tab:stage-trajectory}
\centering
\small
\begin{tabular}{lrrrr}
\toprule
Stage & 4B ASR & 4B Align & 9B ASR & 9B Align \\
\midrule
SFT & $45.3$ & $0.673$ & $53.1$ & $0.561$ \\
SFT$+$DPO & $63.9$ & $0.613$ & $61.1$ & $0.674$ \\
SFT$+$DPO$+$GRPO (Table~\ref{tab:main}) & $70.2$ & $0.619$ & $61.1$ & $0.678$ \\
\bottomrule
\end{tabular}
\end{table}

\subsection{Removing the factuality tier}
\label{app:margin-only}

The GRPO stage ranks each group of candidate rewrites lexicographically: first by a factuality tier, then by the surrogate margin towards the target model.
All policies below start from the same DPO checkpoint ($570$ steps) and are taken at GRPO step $50$.
The factuality-first configuration reaches $70.2\%$ held-out ASR, compared with $66.5\%$ for the strongest margin-only variant and $63.9\%$ for the DPO start (Table~\ref{tab:factuality-tier}).
AlignScore is unchanged at $0.619$.
The two DEV-selected checkpoints differ in $\beta$ and in the selection rule---margin-only was ranked by attack strength, factuality-first by factuality---so this is not a controlled single-factor change, and we do not read it as a causal effect of the tier alone.

\begin{table}[!htb]
\centering
\small
\caption{Factuality-first GRPO versus the strongest margin-only variant, both from the same DPO start.
Held-out ASR is 4-LLM target ASR (\%); Align is AlignScore against the source document.
The two DEV-selected checkpoints differ in $\beta$ and in the selection rule, so the comparison is not a single-factor ablation.}
\label{tab:factuality-tier}
\begin{tabular}{lcc}
\toprule
Configuration & Held-out ASR & Align \\
\midrule
DPO (start) & 63.93 & 0.613 \\
Factuality-first ($\beta{=}0.05$) & 70.15 & 0.619 \\
Strongest margin-only ($\beta{=}0.02$) & 66.50 & 0.619 \\
\bottomrule
\end{tabular}
\end{table}

\subsection{What the Teacher warm start is worth}
\label{app:native-sft}

Removing the Teacher while keeping everything else isolates what the warm start buys.
\emph{Native-SFT} fine-tunes the same backbone on parallel $S\!\rightarrow\!T$ summaries of the same document from the training split and then runs the identical factuality-ranked GRPO stage; it never sees a Teacher rewrite and never runs DPO.
It reaches macro $41.9$ target ASR $[41.3,42.4]$ and $78.3$ source evasion $[77.9,78.7]$, but at an AlignScore of $0.229$ on CNN/DM against the Teacher's $0.682$ --- it moves the label by rewriting far more destructively, which the threat model of Section~\ref{sec:threat} does not permit; per domain the target ASR is $46.5$ $[45.6,47.5]$ on CNN/DM, $39.2$ $[38.0,40.4]$ on ArXiv, $41.9$ $[40.7,43.1]$ on SAMSum, and $40.0$ $[38.9,41.1]$ on WikiHow, with source evasion $81.0$, $75.2$, $79.8$, and $77.1$.
That is well above the RL baselines of Table~\ref{tab:main}, whose macro intervals $[24.6,25.2]$ and $[23.9,24.6]$ do not come close, and about $19$ points of macro ASR below \method{}-4B $[60.3,61.3]$.
Aligning the RL objective with ours therefore closes about half the gap those baselines leave open ($17.3$ of $36.2$ macro points); the remainder is what the Teacher's rewrites and the preference stage supply.

\FloatBarrier


\section{Cost accounting}
\label{app:sec-cost}

\subsection{List prices and compute}
\label{app:tokens}

\paragraph{What the rewrite adds.} Table~\ref{tab:cost-per-1k} separates the two terms an intermediary pays.
Substituting the open model is where the money is, and what it costs depends on the hardware.
The disguise is not: one rewrite costs \$$0.012$ per thousand summaries on a rented consumer card, about $3\%$ of the \$$0.38$ the Gemma summaries themselves cost.
Priced instead on the datacentre card that writes the summaries, the rewrite is \$$0.33$, between $4\%$ and $14\%$ of the premium rates in the table.
Ratios, not absolute figures, are what survive a change of price list.


\begin{table}[!htb]
\caption{Cost of $1{,}000$ CNN/DM summaries.
API rows apply list rates to the stored document and summary token counts; GPU rows price the measured GPU-hours of Table~\ref{tab:decode-throughput} on the card named, without re-timing on it.
All rates were read on 18 September 2026; it is the ratio rather than the absolute figures that we report.}
\label{tab:cost-per-1k}
\centering
\small
\begin{tabular}{@{}llr@{}}
\toprule
Item & Rate or hardware & \$ / $1$k \\
\midrule
\multicolumn{3}{@{}l}{\emph{Premium tier: what a reseller claiming that model would bill}} \\
Grok 4.6 & list & $2.34$ \\
Gemini 3.1 Pro & list & $3.23$ \\
Claude Sonnet & list & $3.51$ \\
GPT-5.6 Sol & list & $6.22$ \\
Claude Opus 5 & list & $8.78$ \\
\midrule
\multicolumn{3}{@{}l}{\emph{Substitution: one Gemma-4 26B summary in place of a commercial one}} \\
Gemma-4 26B & rented A100, \$$1.99$/h & $0.38$ \\
\midrule
\multicolumn{3}{@{}l}{\emph{Disguise: one \method{}-4B rewrite per summary}} \\
\method{}-4B & rented RTX~5060, \$$0.07$/h & $\mathbf{0.012}$ \\
\method{}-4B & rented A100, \$$1.99$/h & $0.33$ \\
\bottomrule
\end{tabular}
\end{table}

All dollar figures are \emph{inference} costs: each new summary pays the API call, or the open-model summary plus the rewrite.
Adapter training is paid once (Table~\ref{tab:train-gpuh}) and is not amortised into the per-thousand figures.
API rows apply list rates to stored document and summary token counts; rewrite rows take the GPU-hours measured on an H200 (Table~\ref{tab:decode-throughput}) and price them on the card named.
The estimate excludes the attacker's one-time costs --- the Round-1 corpus, operator induction, Teacher data generation, and the $1.33$ GPU-hours of student training in Table~\ref{tab:train-gpuh}.


\begin{table}[!htb]
\caption{One-time adapter training on one H200 NVL, in GPU-hours.
This is setup, not the per-rewrite inference cost of Table~\ref{tab:cost-per-1k}. 9B GRPO continues from that size's DPO adapter; the DPO column is the full DPO job.
The additional training of the Gemma-source adapters is not included (Appendix~\ref{app:gemma-source-targeted}).}
\label{tab:train-gpuh}
\centering
\small
\begin{tabular}{lrrrr}
\toprule
Student & SFT & DPO & GRPO & Sum \\
\midrule
\method{}-4B & $0.43$ & $0.79$ & $0.11$ & $1.33$ \\
\method{}-9B & $0.89$ & $1.31$ & $0.16$ & $2.36$ \\
\bottomrule
\end{tabular}
\end{table}

\paragraph{GPU time.} Table~\ref{tab:teacher-crossdomain-cost} gives the GPU hours behind the cost analysis above. Teacher and students were timed on the same hardware at the default decoding settings (students: beam $2$, batch $16$), not at a saturated batch.
Table~\ref{tab:decode-throughput} shows how much cheaper the student becomes when decoding is tuned for throughput, so the cost estimate above is conservative for the attacker.
Batch $1024$ requires greedy decoding; beam $2$ runs out of memory beyond batch $512$.


\begin{table}[!htb]
\caption{\method{}-4B decode throughput on one H200 ($n{=}2397$).
Greedy batch $1024$ is the cheapest student setting that fit in $140$~GB.}
\label{tab:decode-throughput}
\centering
\small
\begin{tabular}{lrr}
\toprule
Setting & ms/item & GPU-h / $2397$ \\
\midrule
Default (beam $2$, batch $16$) & $602$ & $0.40$ \\
Beam $2$, batch $512$ (max that fits) & $370$ & $0.25$ \\
Greedy, batch $1024$ (max that fits) & $114$ & $0.076$ \\
\bottomrule
\end{tabular}
\end{table}


\begin{table}[!htb]
\caption{\method{} rewrite GPU hours, timed on the same hardware at the default decoding settings (students: beam $2$, batch $16$).
Teacher hours cover candidate generation and surrogate scoring with the frozen CNN/DM-built bank ($N{=}5$, $k{=}5$ on CNN/DM; $k{=}2$ on ArXiv).
Student hours are a timed decode of the same $2397$ CNN/DM instances, including model load.}
\label{tab:teacher-crossdomain-cost}
\centering
\small
\begin{tabular}{llr}
\toprule
Method & Domain & GPU-h \\
\midrule
Teacher & CNN/DM  & 2.42 \\
        & ArXiv   & 3.92 \\
        & SAMSum  & 1.16 \\
        & WikiHow & 2.32 \\
\method{}-4B & CNN/DM & 0.40 \\
\method{}-9B & CNN/DM & 0.53 \\
\bottomrule
\end{tabular}
\end{table}

\FloatBarrier


\section{Gemma as the source: open-model impersonation}
\label{app:sec-gemma-source}

This appendix supports Section~\ref{sec:impersonation}.
Gemma-4 26B writes the CNN/DM summaries and the attack must send each one to Gemini, Claude, Grok, or GPT.
All results use the Gemma-5 suite, so that the true source is a labelled class and source evasion is defined; they are not on the 4-LLM scale of Table~\ref{tab:main}.
The realistic case is the one in which the attacker trains for its own open-source writer.

\subsection{Targeted training on a Gemma source}
\label{app:gemma-source-targeted}

The attacker runs the same Teacher on Gemma$\rightarrow\{\textrm{Gemini},\textrm{Claude},\textrm{Grok},\textrm{GPT}\}$ and continues training \method{}-4B on those pairs by DPO, starting from its factuality-ranked GRPO checkpoint.
The result, written \method{}-4B (Gemma-source), is therefore not the \method{}-4B checkpoint of Table~\ref{tab:main}.
It keeps most of what it was trained on before: under the 4-LLM protocol of Table~\ref{tab:main} it still reaches $65.4$ target ASR on the original twelve paths, against the main student's $70.2$.

\emph{Protocol.} CNN/DM Round-2, $200$ documents $\times$ $4$ paths ($n{=}800$), Gemma-5 $\mathrm{Mean}_4$.
These percentages are not on the 4-LLM scale of Table~\ref{tab:main} and are not comparable with the main comparison.
Published baselines are regenerated from Gemma summaries with the same Gemma-4 rewriter as Table~\ref{tab:main}'s prompted methods.
The DPO hyperparameters ($\mathrm{lr}{=}2{\times}10^{-6}$, $\beta{=}0.05$) reuse the $9$B recipe's GRPO rate and were not tuned on this task; a learning rate of $5{\times}10^{-6}$ gives $68.5$ against $68.3$.

Both students exceed the Teacher in this, the realistic, case. The headline rates are those of Table~\ref{tab:impersonation}.
Impersonation is not free on faithfulness---AlignScore $0.550$ is below 5-shot FC ($0.636$), the Teacher ($0.631$), and the unmodified Gemma summary ($0.674$), and PPL is $1.6\times$ that of the Gemma source.


\begin{figure}[!htb]
\centering
\includegraphics[width=\linewidth]{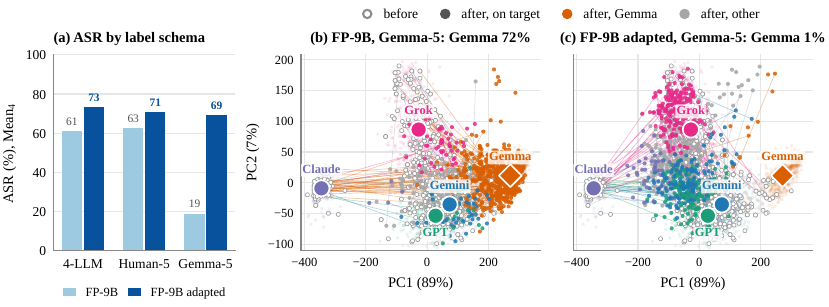}
\caption{Further training prevents \method{}-9B rewrites from being classified as
Gemma.
\textbf{(a)} Adding Gemma as a fifth label reduces the original
\method{}-9B target ASR from $61\%$ to $19\%$, while the further-trained
student retains $69\%$ ASR.
\textbf{(b, c)} Under the Gemma-5 evaluator, many rewrites from the
original student are classified as Gemma ($72\%$), compared with only
$1\%$ after further training.
Hollow points show summaries before rewriting and colored points show
their rewrites.
The additional training uses Gemma-source paths but never a
Gemma-inclusive surrogate.}
\label{fig:adapted}
\end{figure}

\paragraph{The adapted student on the original twelve paths.} Gemma-source training does not cost the student its original ability.
Scored back on the twelve LLM paths under the 4-LLM suite of Table~\ref{tab:main}, the Gemma-source 9B adapter reaches $73.49$ $\mathrm{Mean}_4$, above the backbone control ($61.14$); this is the adapted student shown in Figure~\ref{fig:adapted}.
It starts from the locked \method{}-9B and is trained by DPO on the Gemma-source four-path Teacher rewrites together with the original twelve-path pairs; the extra on-manifold Gemma-source data teach the student to leave the open-model bucket without losing target ASR.
Because it consumes additional training data under a different recipe, it is reported separately from the main comparison.
Table~\ref{tab:per-evaluator-gsrc9b} splits both adapters by evaluator on that suite and on the four-path Gemma-5 suite of Section~\ref{sec:impersonation}.
On twelve paths, RoBERTa is the most permissive evaluator for both students ($85.19$ vs.\ $67.75$) and DeBERTa the strictest.
The $+12.4$ Mean$_4$ lift is uneven: $+17.4$ on RoBERTa against $+7.2$ on DeBERTa.
On the four Gemma paths the unadapted backbone control collapses ($14.41$), and the most permissive evaluator is TF-IDF ($31.00$), not RoBERTa.
The Gemma-source 9B recovers $66.69$ on that same four-path suite, with RoBERTa again the easiest evaluator ($75.25$).


\begin{table}[!htb]
\caption{Target ASR (\%) by evaluator for the \method{}-9B backbone control and the Gemma-source 9B student.
Top: CNN/DM, $12$ LLM paths, $n{=}2397$, domain-matched 4-LLM.
Bottom: Gemma-authored CNN/DM summaries, four Gemma$\to$LLM paths, $n{=}800$, Gemma-5 $\mathrm{Mean}_4$.
The two rows are different adapters.
Bold: most permissive evaluator in that row.}
\label{tab:per-evaluator-gsrc9b}
\centering
\small
\begin{tabular}{lrrrrr}
\toprule
Method & RoBERTa & DeBERTa & GPT-2 & TF-IDF & $\mathrm{Mean}_4$ \\
\midrule
\multicolumn{6}{l}{\emph{12-path 4-LLM}} \\
\method{}-9B & \textbf{67.75} & 57.78 & 61.03 & 57.99 & 61.14 \\
\method{}-9B (Gemma-source) & \textbf{85.19} & 64.96 & 75.76 & 68.04 & 73.49 \\
\midrule
\multicolumn{6}{l}{\emph{Gemma$\to$LLM 4-path, Gemma-5}} \\
\method{}-9B & 11.50 & 6.88 & 8.25 & \textbf{31.00} & 14.41 \\
\method{}-9B (Gemma-source) & \textbf{75.25} & 65.38 & 63.88 & 62.25 & 66.69 \\
\bottomrule
\end{tabular}
\end{table}

\FloatBarrier


\section{Mechanism analyses}
\label{app:sec-mechanism}

The analyses in this appendix use different scoring protocols from Table~\ref{tab:main}.
Absolute values are comparable with that table only where the protocol is 4-LLM on the twelve LLM paths; elsewhere only within-analysis differences are interpreted.

\subsection{Directed-path and domain effects}
\label{app:path-domain}

\begin{figure}[!htb]
\centering
\includegraphics[width=\linewidth]{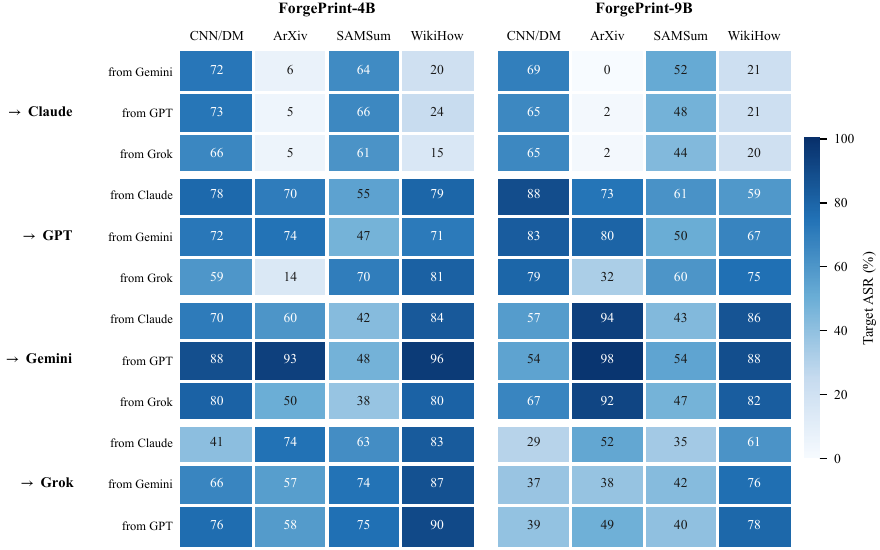}
\caption{Target ASR for each directed path (rows, grouped by target) in each domain (columns), for both students, under domain-matched 4-LLM $\mathrm{Mean}_4$.
Difficulty tracks the target more than the domain, but cell means also show a large path$\times$domain interaction.
The main-text heatmap (Figure~\ref{fig:target-domain}) averages the three sources of each target.}
\label{fig:path-domain}
\end{figure}

Figure~\ref{fig:path-domain} gives the full $12$-path $\times$ $4$-domain matrix behind Figure~\ref{fig:target-domain} for \method{}-4B.
An unweighted sum-of-squares decomposition over its $48$ cell means assigns $32.6\%$ to the directed-path main effect, $14.0\%$ to the domain main effect, and $53.4\%$ to their interaction residual.
Nested inside the path term, the target accounts for $82.6\%$ of path SS, with source-within-target the remainder.
This is a descriptive cell-mean decomposition, not a nested ANOVA or causal estimate.


\subsection{Label-set counterfactual: swapping the fifth class}
\label{app:label-geometry}

This analysis asks one question: if every rewrite is held byte-identical, does the defender's unused fifth class change measured target ASR?
The design is intentionally asymmetric: we fix the attacker and change only the defender.
It is not a matched five-class attack.

The attack is the twelve LLM$\to$LLM paths of Table~\ref{tab:main}.
Three five-class $\mathrm{Mean}_4$ suites on CNN/DM ($n{=}2397$) score the same saved rewrites and differ in one slot: Human, Gemma, or Qwen3.5-9B.

\begin{table}[!htb]
\caption{CNN/DM target ASR (\%) for the same saved rewrites under three fifth-class suites.
Human-5, Gemma-5, and Qwen-5 differ only in the unused fifth label.
The adapted $9$B student is the only policy trained after those classes were introduced.
This is a defender-side counterfactual, not a matched five-class attack.}
\label{tab:label-geometry}
\label{tab:mean5}
\centering
\small
\begin{tabular}{@{}lccc@{}}
\toprule
System & Human-5 ASR & Gemma-5 ASR & Qwen-5 ASR \\
\midrule
Teacher & 52.68 & 27.36 & 31.2 \\
\method{}-4B & 70.36 & 41.39 & 48.0 \\
\method{}-9B & 62.80 & 18.99 & 29.0 \\
\method{}-9B (adapted) & 70.9 & 69.3 & 71.3 \\
\bottomrule
\end{tabular}
\end{table}

Changing the fifth class substantially changes measured target ASR.
Relative to Human-5, Gemma-5 and Qwen-5 drop Teacher / \method{}-4B / \method{}-9B by $25$--$44$ points, while source evasion rises because the new class absorbs mass from the intended target.
Human-5 is nearly inert: RoBERTa assigns none of the students' CNN/DM rewrites to Human, and methods that never query a surrogate move by at most $1.6$ $\mathrm{Mean}_4$ points.
The absorbing class is not the attacker's rewriter.
A Gemma rewriter is absorbed by the Qwen class almost as readily as by the Gemma class ($61.4$ against $75.0$), and a Qwen rewriter is absorbed by both ($63.7$ and $65.6$).

After source-specific post-training, the adapted $9$B student no longer falls into that extra open-model bucket.
It holds $\mathrm{Mean}_4$ target ASR at $73.5$, $70.9$, $69.3$, and $71.3$ under 4-LLM, Human-5, Gemma-5, and Qwen-5, and cuts the absorbed share to $0.8\%$ and $4.0\%$ (Figure~\ref{fig:adapted}, Appendix~\ref{app:gemma-source-targeted}).

\FloatBarrier

\section{Case studies and failure analysis}
\label{app:cases}

This appendix prints two whole rewrites.
Appendix~\ref{app:examples} showed what the operators do on one path; here each case sets one source summary beside the summary the target model itself wrote for the same document and beside the rewrites that two or three systems produced from the source, with the verdict of every held-out evaluator.
The target's own summary is the reference a reader can hold the rewrites against; the evaluators never see it during the attack.
Case~1 shows transfer that the strongest published baseline does not achieve (Appendix~\ref{app:cases-success}); Case~2 shows a failure that lands on a third model rather than the source (Appendix~\ref{app:cases-failure}).
Neither case was chosen by reading the text: each was drawn at random, seed~$42$, from a pool defined in that subsection.
Orange marks the spans the text refers to.
The line after each label gives the label that each of the four domain-matched 4-LLM evaluators assigns to that text, in the order RoBERTa, DeBERTa, GPT-2, TF-IDF, in bold when it is the target.

\subsection{Transfer the baseline does not achieve}
\label{app:cases-success}

Case~1 is a CNN/DM success.
An instance enters the pool when five conditions hold.
The student's rewrite is attributed to $T$ by the RoBERTa evaluator.
The rewrite of augmented zero-shot, the strongest published baseline of Table~\ref{tab:main}, is not.
The student's probability on $T$ lies between the 30th and 70th percentiles of its CNN/DM successes, so the case is a typical success and not its best one.
Its AlignScore is at or above the median of those successes.
The rewrite is one of the $720$ judged CNN/DM rewrites of Appendix~\ref{app:judge} ($60$ documents $\times$ $12$ paths), and the judge labels it \textsc{supported} or \textsc{minor}.
RoBERTa selects the pool; the case reports all four evaluators.
The middle band is not a weak band: the student's probability on $T$ exceeds $0.99$ at both ends of it.

Case~1 moves a Grok summary to Gemini.
Gemini's own summary of the article is plain prose with no evaluative adjectives; three evaluators attribute it to Gemini and DeBERTa to Grok.
The student turns two long sentences into four short ones with one point each, opens with a framing sentence, and drops the free-kick distance and the adjective \emph{lively}.
The judge records both omissions and passes the rewrite.
Three evaluators name Gemini.
Augmented zero-shot adds exclamations and evaluative adjectives, keeps the trailing brace of its completion template (Appendix~\ref{app:baselines}), and is read as GPT or Grok.
The Teacher keeps the sentence shape of the source and stays at Grok for three evaluators.

\begin{casecard}{Case 1 --- CNN/DM, Grok $\rightarrow$ Gemini (\texttt{cnndm\_Corpus~B:22:Grok\_to\_Gemini})}\label{case:cnndm}
\roleline{cUnmod}{\textsc{source summary}}{\verdictline{Grok / Grok / Grok / Grok}}
Manchester United will trial MK Dons teenagers Luke Tingey, an 18-year-old defender known for a viral 40-yard free-kick, and Kyran Wiltshire, a lively 18-year-old midfielder. The pair, part of MK Dons' Youth Alliance South Cup-winning U18 side, will train at Carrington ahead of United's UEFA U19 Youth League campaign.
\roleline{cTarget}{\textsc{target's own summary}}{\verdictline{\textbf{Gemini} / Grok / \textbf{Gemini} / \textbf{Gemini}}}
Manchester United will give trials to MK Dons teenagers Luke Tingey and Kyran Wiltshire. Tingey, a central defender, gained attention for a long-range free-kick, while midfielder Wiltshire has been part of MK Dons' first-team friendlies. Both players, whose U18 team are Youth Alliance South Cup champions, will be assessed at Manchester United's Carrington training ground to potentially strengthen their squad for the UEFA U19 Youth League.
\roleline{cBase}{\textsc{aug.\ zero-shot}}{\verdictline{GPT / \textbf{Gemini} / Grok / Grok}}
\cue{Exciting news for football fans!} Manchester United is set to welcome two rising stars, Luke Tingey and Kyran Wiltshire, to Carrington for a special trial. Tingey, an 18-year-old defender who has already captured attention with a viral 40-yard free-kick, joins Wiltshire, a dynamic 18-year-old midfielder. Both players bring impressive pedigrees from MK Dons' trophy-winning U18 side, and they will be honing their skills ahead of United's upcoming UEFA U19 Youth League campaign. \cue{It's a fantastic opportunity to witness the next generation of talent in action! \}}
\roleline{cTeach}{\textsc{teacher}}{\verdictline{Grok / \textbf{Gemini} / Grok / Grok}}
Manchester United will trial two MK Dons teenagers, Luke Tingey and Kyran Wiltshire, ahead of the club's UEFA U19 Youth League campaign. Tingey is an 18-year-old defender known for a viral 40-yard free-kick, and Wiltshire is an 18-year-old midfielder. Both players have been part of the MK Dons U18 side that won the Youth Alliance South Cup and are expected to train at Carrington.
\roleline{cOurs}{\textsc{ForgePrint-4B}}{\verdictline{\textbf{Gemini} / \textbf{Gemini} / \textbf{Gemini} / Grok}}
\cue{Manchester United is preparing for its upcoming UEFA U19 Youth League campaign by conducting trials of promising young players.} \cue{Specifically,} the club will assess 18-year-old defender Luke Tingey and 18-year-old midfielder Kyran Wiltshire. Both players are members of the MK Dons U18 squad that recently won the Youth Alliance South Cup. \cue{This trial period will involve} training sessions at Carrington.
\end{casecard}

\subsection{Where the student fails on ArXiv}
\label{app:cases-failure}

ArXiv is the student's weakest domain (Table~\ref{tab:main}), and the target decides the failures.
Under the RoBERTa evaluator the student misses Claude as the target on 579 of 600 ArXiv instances, against 147 of 600 when Claude is the source; Figure~\ref{fig:path-domain} gives the four-evaluator view, $4.6$--$6.5\%$ on the three paths into Claude.
Of the 1167 ArXiv misses over all twelve paths, 516 stay at the source and 651 land on a third model.
These counts are one evaluator's argmax and are not on Table~\ref{tab:main}'s scale.
Case~2 shows a miss that lands on a third model.
It was drawn at random, seed~$42$, from the Claude-involving ArXiv misses of that kind whose top probability lies in the middle band of that class.
There is no LLM judge on ArXiv; the target's own summary is capped at $220$ words so that the case fits a page.

Case~2 aims a Claude summary at Gemini and lands on GPT for three evaluators and Grok for the fourth.
The student halves the length, removes every heading and bullet, and writes one dense paragraph opened by participles and closed by \emph{consequently}: the same GPT register as a successful GPT-targeted rewrite.
Gemini's own summary of the paper is three prose paragraphs that open with \emph{This paper investigates}.
The Teacher makes a lighter edit, keeps that same opening sentence from the source, adds a reporting frame, and is read as Gemini by all four.
Suppressing the source is not the same as reaching the target.
This is the pattern behind the miss destinations of Section~\ref{sec:two-criteria}: the rewrite counts as evasion and fails as transfer.

\begin{casecard}{Case 2 --- ArXiv, Claude $\rightarrow$ Gemini (\texttt{arxiv\_Corpus~B:60:Claude\_to\_Gemini})}\label{case:fail-b}
\roleline{cUnmod}{\textsc{source summary}}{\verdictline{Claude / Claude / Claude / Claude}}
\# Summary\par\smallskip
This paper investigates the off-shell propagating behavior of photons in atoms during photon-atom scattering using bound-state quantum electrodynamics (QED).\par\smallskip
\#\# Main Points:\par\smallskip
\hangindent=1.0em\hangafter=1 1. \textbf{Problem Statement}: When photons propagate through optical media, they become ``off-shell'' (violating the relativistic energy-momentum relation), unlike in vacuum where they are ``on-shell.'' The paper aims to explain this off-shell behavior during photon-atom interactions.\par\smallskip
\hangindent=1.0em\hangafter=1 2. \textbf{Limitations of Existing Models}: Standard quantum optics models (Jaynes-Cummings, semi-classical approaches) rely on the electric dipole approximation, which restricts them to long-wavelength light and cannot describe photon propagation within atoms.\par\smallskip
\hangindent=1.0em\hangafter=1 3. \textbf{Proposed Solution}: Bound-state QED with S-matrix formalism is more suitable because it:\newline
\hspace{1.2em}\hangindent=2.2em\hangafter=1 -- Handles light-atom interactions at all wavelengths\newline
\hspace{1.2em}\hangindent=2.2em\hangafter=1 -- Naturally describes off-shell photon propagation through Feynman propagators\newline
\hspace{1.2em}\hangindent=2.2em\hangafter=1 -- Avoids the electric dipole approximation limitations\par\smallskip
\hangindent=1.0em\hangafter=1 4. \textbf{Key Findings}:\newline
\hspace{1.2em}\hangindent=2.2em\hangafter=1 -- Only ground-state bound electrons serve as initial/final S-matrix states (excited states decay via spontaneous emission)\newline
\hspace{1.2em}\hangindent=2.2em\hangafter=1 -- Feynman propagators of bound electrons determine photon propagation properties in atoms\newline
\hspace{1.2em}\hangindent=2.2em\hangafter=1 -- These propagators exhibit superluminal but causality-preserving amplitudes\newline
\hspace{1.2em}\hangindent=2.2em\hangafter=1 -- The framework explains various scattering processes: Rayleigh, Compton, and Raman scattering\par\smallskip
\hangindent=1.0em\hangafter=1 5. \textbf{Experimental Implications}: The approach is particularly valuable for studying short-wavelength light (X-rays, gamma rays) interactions with atoms, where traditional approximations fail.
\roleline{cTarget}{\textsc{target's own summary}}{\verdictline{\textbf{Gemini} / \textbf{Gemini} / \textbf{Gemini} / \textbf{Gemini}}}
\cue{This paper investigates} the ``off-shell'' behavior of photons propagating within atoms during photon-atom scattering processes, using the framework of bound-state quantum electrodynamics (QED). Traditional quantum optics models, relying on the electric dipole approximation, are limited to long-wavelength light and cannot accurately describe photon propagation within atoms. Bound-state QED, however, offers a more fundamental approach that is valid for all wavelengths and naturally incorporates the off-shell behavior of photons.\par\smallskip
The study utilizes the S-matrix formalism of bound-state QED to analyze photon-atom scattering. It identifies that only ground states of bound electrons are suitable as initial and final states for the S-matrix due to spontaneous emission. Various photon-atom scattering processes, including Rayleigh, Compton, and Raman scattering, are explained through specific Feynman diagrams within this framework. The internal electron lines in these diagrams, represented by Feynman propagators, phenomenally describe the off-shell amplitudes of photons within atoms. These propagators exhibit a superluminal yet causal property, which is linked to the observed superluminal group velocity of light pulses in atomic media.\par\smallskip
The paper also establishes a connection between the bound-state QED approach and the widely used light-atom interaction models in quantum optics, showing that the latter can be viewed as an electric dipole approximation of the former. While quantum optics models are suitable for time-differential processes, bound-state QED with its S-matrix formalism is better suited for time-integrating processes like photon-atom scattering, especially for short-wavelength light where the electric dipole approximation fails. The research suggests that experimental observations of off-shell photon propagation, particularly with X-ray photons, are possible with advanced light sources and experimental setups.
\roleline{cTeach}{\textsc{teacher}}{\verdictline{\textbf{Gemini} / \textbf{Gemini} / \textbf{Gemini} / \textbf{Gemini}}}
\cue{This paper investigates} the off-shell propagating behavior of photons in atoms during photon-atom scattering using bound-state quantum electrodynamics (QED). While photons in a vacuum are ``on-shell,'' they become ``off-shell'' when propagating through optical media, a phenomenon this research aims to explain. \cue{The study notes that} standard quantum optics models, such as the Jaynes-Cummings model, are limited by the electric dipole approximation and cannot describe photon propagation within atoms. Instead, the research utilizes bound-state QED with S-matrix formalism, which handles light-atom interactions at all wavelengths and naturally describes off-shell propagation through Feynman propagators. Key findings indicate that Feynman propagators of bound electrons determine photon propagation properties, exhibiting superluminal but causality-preserving amplitudes. This framework successfully explains Rayleigh, Compton, and Raman scattering, offering significant value for studying short-wavelength interactions like X-rays and gamma rays.
\roleline{cOurs}{\textsc{ForgePrint-4B}}{\verdictline{GPT / GPT / GPT / Grok}}
This research investigates the off-shell propagation of photons within atoms during photon-atom scattering, \cue{utilizing bound-state quantum electrodynamics (QED)} to overcome the limitations of standard Jaynes-Cummings and semi-classical models. \cue{By employing S-matrix formalism,} the framework avoids the electric dipole approximation and naturally handles light-atom interactions at all wavelengths. Although excited states decay via spontaneous emission, only ground-state bound electrons serve as the initial and final S-matrix states, with their Feynman propagators determining the properties of photon propagation. These propagators exhibit superluminal amplitudes that remain causality-preserving. \cue{Consequently,} the framework explains Rayleigh, Compton, and Raman scattering, offering a solution for the study of short-wavelength light interactions with atoms.
\end{casecard}

\section{Potential limitations}
\label{app:limitations}

This appendix outlines several limitations of the current study and the
scope in which the results should be interpreted.

\begin{itemize}

\item \textbf{task and model scope.}
Our experiments focus on controlled summarisation and closed-set attribution
among four commercial models.
This setting allows different models to express the same underlying content,
which helps separate changes in writing style from changes in topic.
We do not test whether the same results extend to open-ended generation,
substantially larger label spaces, or other model tiers.

\item \textbf{defender setting.}
Our conclusions concern supervised text-only attribution classifiers trained
on unmodified text.
The attack never queries these evaluators and does not use their parameters
or training data.
We do not evaluate adaptive classifiers retrained on targeted rewrites, or
other provenance mechanisms such as watermarks and execution records, which
operate under different assumptions
(Section~\ref{sec:threat}).

\item \textbf{training variability.}
Each Student configuration is trained once.
The confidence intervals in Table~\ref{tab:main} therefore measure variation
over test documents rather than variation across independent training runs.
All main comparisons use the same frozen evaluation protocol.

\item \textbf{rewrite quality.}
Rewrite quality is evaluated with AlignScore, MiniCheck, and a separate LLM
judge on a subset of CNN/DM
(Appendix~\ref{app:judge}).
We do not include human evaluation, so these results should be interpreted as
metric-based assessments of faithfulness rather than human judgments of
overall rewrite quality.
We also analyze rewrite length and added content, but do not claim to fully
separate every surface feature from model-specific style.

\item \textbf{cross-domain operator transfer.}
The operator bank is built once on CNN/DM and then reused unchanged on ArXiv,
SAMSum, and WikiHow.
The cross-domain results therefore measure transfer of a fixed bank; they do
not show how much additional gain a domain-specific bank might provide.

\item \textbf{cost estimates.}
Appendix~\ref{app:tokens} reports marginal inference costs using published
prices and measured decoding time.
These estimates exclude one-time data collection and training costs and should
not be interpreted as the operating cost of any model provider.

\end{itemize}

\end{document}